\documentclass[10pt]{IEEEtran}

\usepackage{afterpage}
\usepackage{algorithmicx}
\usepackage[linesnumbered,ruled,longend]{algorithm2e}
\SetKwInOut{Input}{Input}
\SetKwInOut{Output}{Output}

\SetKwProg{Fn}{Function}{\string:}{end}

\usepackage{lipsum}
\usepackage{amsfonts} 
\usepackage{amsmath}
\usepackage{amssymb}
\usepackage{array} 

\usepackage{boxedminipage}
\usepackage{caption} 
\usepackage{cite}
\usepackage{color} 
\usepackage{enumerate}
\usepackage{float}
\usepackage{framed}

\usepackage{graphics}
\usepackage{graphicx}
\usepackage{indentfirst}
\usepackage{latexsym}
\usepackage{mathtools}
\usepackage{multirow} 
\usepackage{booktabs}
\usepackage{picinpar}
\usepackage{placeins}
\usepackage{subfigure}
\usepackage{theorem} 
\usepackage[para]{threeparttable}
\usepackage{wrapfig} 
\usepackage{textcomp}
\usepackage{verbatim}
\usepackage{multicol}
\usepackage{stfloats}
\usepackage{soul}
\include{macros}
\usepackage{bbold}

\begin{document}

\title{Graph Neural Network Aided Deep Reinforcement Learning for Resource Allocation in Dynamic Terahertz UAV Networks}

\author{Zhifeng~Hu, and Chong~Han,~\IEEEmembership{Senior~Member,~IEEE}
		
		\thanks{
			\par
			Zhifeng Hu is with the Terahertz Wireless Communications (TWC) Laboratory, Shanghai Jiao Tong University, Shanghai 200240, China (email: zhifeng.hu@sjtu.edu.cn).
            \par Chong Han is with the Terahertz Wireless Communications (TWC) Laboratory and also the Cooperative Medianet Innovation Center (CMIC), School of Information Science and Electronic Engineering, Shanghai Jiao Tong University, China (Email:
chong.han@sjtu.edu.cn).
		}
}

\maketitle

\begin{abstract}
Terahertz (THz) unmanned aerial vehicle (UAV) networks with flexible topologies and ultra-high data rates are expected to empower numerous applications in security surveillance, disaster response, and environmental monitoring, among others.
However, the dynamic topologies hinder the efficient long-term joint power and antenna array resource allocation for THz links among UAVs.
Furthermore, the continuous nature of power and the discrete nature of antennas cause this joint resource allocation problem to be a mixed-integer nonlinear programming (MINLP) problem with non-convexity and NP-hardness.
Inspired by recent rapid advancements in deep reinforcement learning (DRL), a graph neural network (GNN) aided DRL algorithm for resource allocation in the dynamic THz UAV network with an emphasis on self-node features (GLOVE) is proposed in this paper, with the aim of resource efficiency (RE) maximization.
When training the allocation policy for each UAV, GLOVE learns the relationship between this UAV and its neighboring UAVs via GNN, while also emphasizing the important self-node features of this UAV.
In addition, a multi-task structure is leveraged by GLOVE to cooperatively train resource allocation decisions for the power and sub-arrays of all UAVs.
Experimental results illustrate that GLOVE outperforms benchmark schemes in terms of the highest RE and the lowest latency.
Moreover, unlike the benchmark methods with severe packet loss, GLOVE maintains zero packet loss during the entire training process, demonstrating its better robustness under the highly dynamic THz UAV network.
\end{abstract}

\section{Introduction}
\label{sec: intro}

As airborne devices, unmanned aerial vehicles (UAVs) are capable of supporting a plethora of promising missions from the sky~\cite{yan2019comprehensive}, including target surveillance, environmental monitoring, as well as disaster response, etc.
Unlike traditional manned aircraft, UAVs can satisfy sensing and communication demands in harsh or hostile regions without the risk of endangering personnel.
Fortunately, the tremendous development in low-cost UAV fabrication has paved the way for the practical implementation of affordable UAV systems~\cite{azari2022thz}.
In addition, owing to the emerging low Earth orbit (LEO) mega-constellations like Starlink, LEO satellites can be harnessed to enable global remote control for UAVs and relay data from UAVs in traditionally inaccessible areas~\cite{kokkoniemi2021channel}.

In comparison with the conventional single-UAV systems, UAV networks benefit from higher scalability and faster task completion via the deployment of multiple UAVs in a mission region~\cite{gupta2015survey}.
Additionally, different from the terrestrial networks with fixed base stations, the high maneuverability of UAVs enables flexible reconfigurability of networks~\cite{zeng2017energy}.
Moreover, compared to high-altitude platforms (HAPs), UAV networks exhibit better channel conditions in links with terrestrial devices, due to shorter transmission distances~\cite{zeng2017energy}.
Foreseeing the two trends of a large quantity of connected devices and quest for ultra-fast data rates, Terahertz (THz) band (0.1-10 THz) communications are deemed as a prospective technology to empower ultra-high-rate transmissions for UAV networks owing to the ultra-broad bandwidth~\cite{azari2022thz}.
There are several advantages that make THz UAV networks attractive.
First, the ultra-high-capacity THz links are able to serve the missions of UAVs that require large traffic volumes with low-latency experience~\cite{akyildiz2022terahertz}. 
Second, due to the unobstructed THz links among UAVs and large-scale arrays of sub-millimeter-long antennas, line-of-sight (LoS) multiple-input and multiple-output (MIMO) and hybrid beamforming can be leveraged to extend propagation distance, mitigate interference, and enhance spectral efficiency~\cite{han2021hybrid,azari2022thz}.
Last but not least, the ultra-broadband THz waves can provide ultra-high-resolution sensing for the surveillance and monitoring missions of UAV networks.
Hence, the adoption of THz communications can empower the integration of THz sensing and communication (THz ISAC) in UAV networks~\cite{meng2023uav}, which can reduce hardware costs and further improve spectral efficiency.

Despite these advantages, THz UAV networks still face significant challenges.
First, in THz systems, the efficient management of sub-arrays for hybrid beamforming and the costs caused by transmit power consumption are both key concerns~\cite{hu2023deep,zhai2019antenna}.
Therefore, to enable resource-efficient THz communications in UAV networks, it is essential to minimize the usage of power and sub-arrays.
Second, different from fixed terrestrial backhaul networks as well as satellite networks with predefined satellite trajectories, the topologies of the UAV network might be highly dynamic and unpredictable, due to the needs of some missions (e.g., the fickle movement of surveilled targets~\cite{zhang2024cooperative}). 
Thus, specific designs for the long-term resource allocation strategy in the dynamic THz UAV network are motivated.
Third, resulting from the continuous property of the power and the discrete property of sub-arrays, the long-term joint power and sub-array allocation is a non-convex NP-hard mixed-integer nonlinear programming (MINLP) problem~\cite{jeong2022transport}.

Recently, some research attempts have commenced on resource allocation optimization in UAV networks with the aim of resource efficiency (RE, referring to the data rate per resource usage) maximization.
Several works maximize RE of the UAV network in a non-learning manner~\cite{tong2022joint,kim2020energy}.
More specifically, they optimize the resource allocation on the basis of iterative optimization methods (e.g., iterative Lagrange multiplier method).
However, in the dynamic THz UAV network, the iterative calculation in every time slot can incur prohibitively high computational complexity, especially for the NP-hard MINLP problem.
In addition, the optimization results of these works rely on the exact prediction of traffic demands for all devices in all time slots, which is impractical in highly dynamic scenarios.
Consequently, existing non-learning solutions are not viable for resource allocation optimization in the dynamic THz UAV network.

Thanks to the recent progress of machine learning, reinforcement learning (RL) and deep reinforcement learning (DRL) approaches~\cite{cui2019multi,zhou2022resource,bai2025dynamic,park2023joint} are proposed as well, which can optimize long-term resource allocation strategies without multiple iterations per time slot.
The RL method~\cite{cui2019multi} optimizes resource allocation by deploying a lookup table, which requires all pairs of resource allocation action and environment state of the wireless network to be visited many times.
Hence, the RL method is infeasible in complex UAV networks, where the state and action spaces are extremely large or even continuous.
In contrast, DRL algorithms~\cite{zhou2022resource,bai2025dynamic,park2023joint} can utilize deep neural networks to learn the underlying relationship between the resource allocation policy and the environment state.
As a result, DRL methods can infer optimal allocation actions even under unvisited states in complicated scenarios.
Nevertheless, these DRL schemes focus on communications between UAVs and terrestrial users while neglecting resource allocation for links among UAVs.
Moreover, these existing works, including both learning and non-learning approaches, overlook the sub-array management, which is important for the THz transceivers with massive antenna arrays.

To address the above mentioned challenges, we propose a graph neural network (GNN) aided DRL algorithm for resource allocation in the dynamic THz UAV network with an emphasis on self-node features (GLOVE), with the target of long-term RE maximization.
Specifically, we focus on the resource-efficient joint power and sub-array allocation of the THz links within the UAV network that sends the collected data from surveillance missions or terrestrial devices to the LEO satellite.
In particular, the proposed DRL algorithm can jointly determine the transmitting and receiving sub-arrays, as well as the transmit power for THz links among UAVs in a long-term period.
The main contributions of this work are summarized as follows.
\begin{itemize}
    \item In the dynamic THz UAV network, we formulate a joint power and sub-array resource allocation problem.
    This problem aims at RE maximization for THz links among UAVs with the consideration of the peculiarities of THz propagation and beam misalignment for transceivers on UAVs.
    Moreover, the formulated problem further takes into account physical constraints pertaining to the limitations of power and sub-array resources, as well as the alleviation of latency and packet loss.
    \item We propose a DRL framework tailored to the joint power and sub-array resource allocation problem in the THz UAV network.
    Particularly, based on the state of UAVs and THz links, this framework encourages DRL to generate resource allocation actions with the target of long-term reward maximization (i.e., RE maximization) without relying on accurate predictions of traffic demands.
    In addition, the customized definitions of the action and the reward mitigate the violations of the aforementioned physical constraints.
    \item Under the proposed DRL framework, we design a GLOVE algorithm to solve the long-term joint power and sub-array allocation problem.
    In the highly dynamic THz UAV network, when GLOVE generates resource allocation action for every UAV, the tailored architecture not only emphasizes the critical self-node features of this UAV, but also leverages GNN to capture the relationship between this UAV and its adjacent UAVs via graph connectivity structure in each time slot.
    To cooperatively train power assignment and sub-array allocation for all UAVs, the multi-task architecture is adopted in GLOVE as well.
    \item We conduct simulations to evaluate the performance of the proposed GLOVE algorithm.
    Compared to benchmark methods, the GLOVE algorithm realizes the highest RE, while achieving the lowest latency.
    One step further, different from benchmark solutions with a large number of lost packets, GLOVE maximizes long-term RE without packet loss, suggesting its superior robustness against the dynamic topology of the THz UAV network.
    Experimental results also show that the proposed mechanism that puts an emphasis on self-node features can enhance RE of GNN-aided DRL, while alleviating the latency and packet loss, without significantly sacrificing computational efficiency and memory efficiency.
\end{itemize}

The rest of this paper is organized as follows.
In Sec.~\ref{sec: system}, we present the model of the THz UAV network, as well as formulate the long-term joint power and sub-array allocation problem.
To solve the formulated problem, we design a customized DRL framework, which is elaborated in Sec.~\ref{sec: DRL description}.
The proposed GLOVE algorithm is detailed in Sec.~\ref{sec: GNN-DRL}.
Then, we assess and analyze the performance of the proposed algorithm in Sec.~\ref{sec: result}.
Finally, we conclude this paper in Sec.~\ref{conclusion}.


\section{System Model and Problem Formulation}
\label{sec: system}

In this section, a THz UAV network is taken into consideration, which encompasses multiple UAVs that undertake transmission or surveillance tasks.
Moreover, one UAV establishes an aerial-space link with a LEO satellite, serving as the gateway that connects this UAV network with a space-based network. 
THz links are utilized to deliver data from all UAVs to this gateway.
Then, we formulate a long-term joint power and sub-array allocation problem, with the aim of RE maximization in this network.

\subsection{THz UAV Network Model}
\label{sec: uav model}

\begin{figure}[t] 
\centering
        \includegraphics[width=\linewidth]{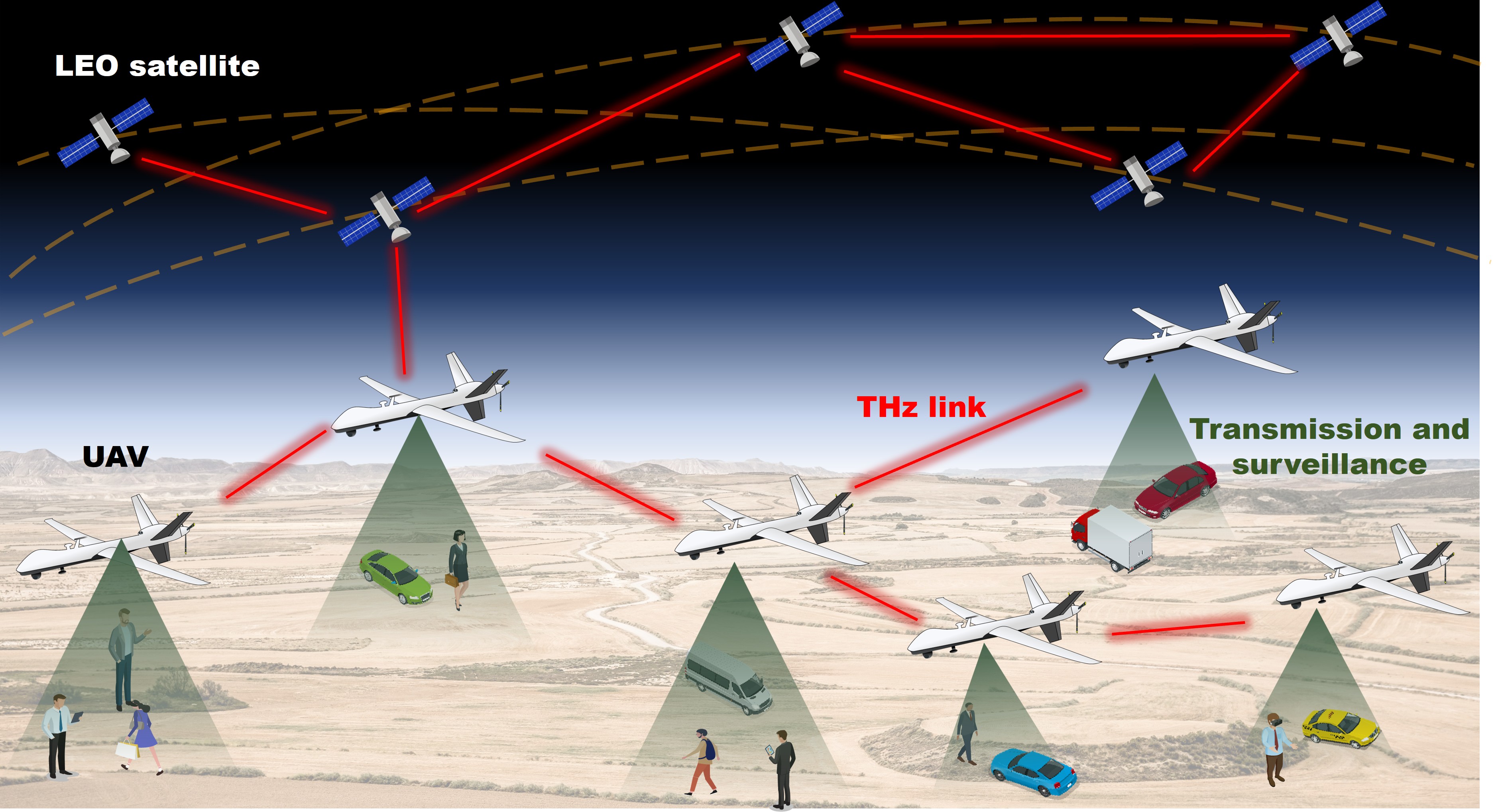} 
        \caption{THz UAV network.} 
        \label{fig: THz uav network}
\end{figure}

As shown in Fig.~\ref{fig: THz uav network}, $N$ UAVs form a THz UAV network in a mission region.
Particularly, each UAV in this network is in charge of surveilling targets or collecting data from devices on the ground.
Among these UAVs, a UAV serves as the header UAV that links the UAV network with the LEO satellite network.
More specifically, the header UAV is the gateway that aggregates the data from all UAVs and transmits the aggregated data to the closest LEO satellite with the assistance of high-gain antennas~\cite{kokkoniemi2021channel}.
As a result, UAVs in this network are capable of transmitting data from hostile, undeveloped, or harsh areas to users across the globe.
For the purpose of high transmission rates, THz links with $K$ sub-bands are utilized to support the links among UAVs.

By considering the mobility, the UAV network deploys a mesh-like architecture, for ease of flexible reconfigurability under dynamic UAV positions.
In particular, each UAV can connect with anyone among the neighboring UAVs in proximity (i.e., within the maximal propagation distance $d_\text{max}$) to transmit its data. 
In addition, each UAV continuously sends probing messages to confirm the link status with the neighboring UAVs~\cite{vestin2017low}.
Consequently, every UAV can rapidly update its routing table to identify the neighboring UAVs that become inaccessible, as well as the currently accessible UAVs.

The collected data in each UAV follows the fractional Brownian motion process, which is a commonly adopted accurate traffic demand model~\cite{fidler2014guide,hu2023deep}.
Moreover, every UAV is equipped with a first-in-first-out (FIFO) buffer for data packet transmissions.
Specifically, the packets in the buffer of each UAV are sent in the order in which they arrive at this buffer.
Furthermore, the storage capacity of the buffer is $\Omega$ data packets. 
Incoming packets are discarded when the buffer is full and cannot accommodate more data packets, resulting in packet loss.

\subsection{THz Transmission Model}
\label{sec: thz transmission model}

The transceivers of the UAV network deploy the dynamic hybrid beamforming structure~\cite{yan2020dynamic} that is tailored for THz communications with single-stream transmissions.
This design provides two-fold benefits.
On one hand, each UAV is allowed to flexibly adjust the number of allocated sub-arrays.
On the other hand, the interference from other links can be alleviated~\cite{cao2007multihop}.
In particular, every sub-array adopts a planar array configuration with $M_x\times M_y$ antennas.
Therefore, for the signal with the wavelength $\lambda$, the elevation angle $\theta$, and the azimuth angle $\phi$, the steering vector of the planar array is computed as
\begin{equation}
\begin{aligned}
    \label{eq:steeringVector}
    \mathbf{a}(\phi, \theta)=&\dfrac{1}{\sqrt{M_x M_y}}\left[1,\dots,e^{\tilde{j}\frac{2\pi d_0}{\lambda}(m_x\sin(\theta)\cos(\phi)+m_y\cos(\theta))},\right.\\
    &\quad \dots,\left.e^{\tilde{j}\frac{2\pi d_0}{\lambda}((M_x-1)\sin(\theta)\cos(\phi)+(M_y-1)\cos(\theta))}\right]^T,
\end{aligned}
\end{equation}
where $0\leq m_x \leq M_x-1$, $0\leq m_y \leq M_y-1$, $d_0$ refers to the spacing between every pair of adjacent antennas, and $\tilde{j}$ denotes the imaginary unit.

Consequently, the MIMO channel response for the $k^\text{th}$ sub-band of the THz link between a UAV $i$ (i.e., the transmitter) with $S_\text{Tx}(i,j)$ transmitting sub-arrays and another UAV $j$ (i.e., the receiver) with $S_\text{Rx}(j,i)$ receiving sub-arrays is given by~\cite{walter2017time, peng2021hybrid}
\begin{equation}
\begin{aligned}
    \label{eq:MIMOresponse}
    \boldsymbol{H}(i,j,k)=&\sqrt{(S_\text{Tx}(i,j)M_x M_y)(S_\text{Rx}(j,i)M_x M_y)}G_\text{Tx}G_\text{Rx}
    \\&\cdot e^{\tilde{j}2\pi f_d \tau_d}\mathbf{a}_\text{Rx}(\phi_\text{Rx},\theta_\text{Rx})\mathbf{a}_\text{Tx}^*(\phi_\text{Tx},\theta_\text{Tx})\alpha(i,j,k) G_{m},
\end{aligned}
\end{equation}
where $G_\text{Tx}$ and $G_\text{Rx}$ stand for the antenna gains for the transmitter and receiver, respectively.
$f_d$ represents the Doppler frequency.
$\tau_d$ symbolizes the delay of this link.
$\mathbf{a}_\text{Tx}$ and $\mathbf{a}_\text{Rx}$ represent the steering vectors for the transmitter and receiver, respectively.
Particularly, $\theta_\text{Tx}/\theta_\text{Rx}$ and $\phi_\text{Tx}/\phi_\text{Rx}$ are the elevation and azimuth angles of departure/arrival, respectively.
$(\cdot)^*$ refers to the conjugate transpose operator.
Furthermore, the mobility of UAVs and environmental disturbances (e.g., wind) might result in beam misalignment.
We adopt the commonly-used model in~\cite{chang2021joint,boulogeorgos2019analytical} for the gain brought by the THz beam misalignment, which is denoted by $G_m$. 
Additionally, for the $k^\text{th}$ sub-band, the path gain $\alpha(i,j,k)$ of the THz transmission between the UAVs $i$ and $j$ with the propagation distance $d(i,j)$ is expressed as
\begin{equation}
\label{eq:pathloss}
  |\alpha(i,j,k)|^2=
  \begin{aligned}
  &\left(\dfrac{c}{4\pi f_k d(i,j)} \right)^2e^{\int_{D_{i}}^{D_{j}}{\left(-g_{abs}(f_k,d')\right)}\mathrm{d} d'},
  \end{aligned}
\end{equation}
where $D_{i}/D_{j}$ represents the three-dimensional position vector of the UAV $i/j$.
$c$ denotes the speed of light.
$f_k$ corresponds to the carrier frequency of the $k^\text{th}$ sub-band, while $g_{abs}(f_k,d')$ represents the medium absorption coefficient of aerial THz propagation for this sub-band~\cite{yang2024universal}.

According to~\eqref{eq:MIMOresponse}, by further accounting for the hybrid beamforming with the analog and digital precoding matrices $\boldsymbol{W_{A}}(i,j,k)$ and $\boldsymbol{W_{D}}(i,j,k)$, as well as the analog and digital combining matrices $\boldsymbol{C_A^*}(j,i,k)$ and $\boldsymbol{C_D^*}(j,i,k)$, the channel response incorporating beamforming is given by
\begin{equation}
\begin{aligned}
   h(i,j,k)=&
    \boldsymbol{C_D^*}(j,i,k)\boldsymbol{C_A^*}(j,i,k)\boldsymbol{H}(i,j,k)\\
    &\cdot \boldsymbol{W_{A}}(i,j,k)\boldsymbol{W_{D}}(i,j,k).
\end{aligned}
\label{eq:SISOresponse} 
\end{equation}
By leveraging the beamforming solution in~\cite{yan2020dynamic}, the optimal precoding and combining matrices, namely, $\boldsymbol{W_{A}}(i,j,k)$, $\boldsymbol{W_{D}}(i,j,k)$, $\boldsymbol{C_A^*}(j,i,k)$, and $\boldsymbol{C_D^*}(j,i,k)$, can be effectively attained.

Based on the channel response in~\eqref{eq:SISOresponse}, the signal-to-interference-plus-noise ratio (SINR) for the $k^\text{th}$ sub-band of the THz link from the UAV $i$ to the UAV $j$ can be obtained as
\begin{equation}
\gamma(i,j,k)=\frac{P(i,j,k)|{h}(i,j,k)|^2}
{I_{s}(i,j,k)+||\boldsymbol{C_D^*}(j,i,k)\boldsymbol{C_A^*}(j,i,k)||_F^2\sigma^2},
\label{eq:gamma}
\end{equation}
where $P(i,j,k)$ denotes the allocated transmit power of the UAV $i$ for the $k^\text{th}$ sub-band.
$\sigma^2$ indicates the noise variance.
$||\cdot||_F$ is the operator of the Frobenius norm.
$I_{s}(i,j,k)$ symbolizes the interference of the received signal with the self-interference cancellation and the aforementioned hybrid beamforming, which can be modeled as a Gaussian variable~\cite{zhang2019joint,lei2020deep}.
For the THz transmission from the UAV $i$ to the UAV $j$, the channel capacity is the summation of the capacities for all sub-bands,  which is expressed as
\begin{equation}
	\label{eq:rate}
R(i,j)=\sum\limits_{k=1}^{K}B\log_2(1+\gamma(i,j,k)),
\end{equation}
where $B$ represents the bandwidth for sub-bands.

\subsection{Joint Power and Sub-array Allocation Problem Formulation}
\label{sec: problem formulation}

This work aims to optimize the long-term resource allocation policy $\pi_p$ in the entire THz UAV network, which governs the resource allocation decisions for both power and sub-array resources for all UAVs (symbolized by $\boldsymbol{P}$ and $\boldsymbol{S}$, respectively), with the goal of RE maximization.
For the given traffic demands of the UAV network, RE maximization is tantamount to resource usage minimization.
In this paper, we focus on resource usage minimization with respect to power and sub-arrays, which are two crucial types of resources usually considered in THz networks~\cite{zhai2019antenna,hu2023deep}.
Additionally, to support rapid routing adjustments in response to dynamic UAV positions with the target of RE maximization, the low-complexity resource-efficient routing strategy in~\cite{hu2023deep} is adopted to determine the topology of the THz UAV network.
This routing method can efficiently benefit the resource usage minimization by jointly taking into account the path loss of THz waves as well as the hop count.

The usage ratio of the sub-arrays for the UAV $i$ includes the utilization ratios of both transmitting and receiving sub-arrays, which are formulated as~\eqref{eq: transmitting array occupation} and~\eqref{eq: receiving array occupation}, respectively.
\begin{subequations}
        \label{eq: Tx and Rx array occupation}
        \begin{align}
        \label{eq: transmitting array occupation}
        U_{S,\text{Tx}}(i)&=\sum\limits_{j\in\{1,2,\dots,N\}\backslash\{i\}}\dfrac{S_\text{Tx}(i,j)}{S_\text{max}},\\
        \label{eq: receiving array occupation}
        U_{S,\text{Rx}}(i)&=\sum\limits_{j\in\{1,2,\dots,N\}\backslash\{i\}}\dfrac{S_\text{Rx}(i,j)}{S_\text{max}},
        \end{align}
\end{subequations}
where $\{1,2,\dots,N\}$ is composed of the indices of all UAVs in the THz UAV network. 
$S_\text{max}$ represents the maximal number of sub-arrays that can be provided by a UAV.
For ease of sub-array management and fairness, if a UAV relays data from multiple UAVs, the same number of sub-arrays is utilized to receive data from each UAV.
The overall sub-array occupation ratio is the summation of transmitting and receiving sub-array usage ratios, as
\begin{equation}
    \label{eq: array occupation}
            U_{S}(i)=U_{S,\text{Tx}}(i)+U_{S,\text{Rx}}(i).
\end{equation}

In parallel, the power usage quotient for the UAV $i$ with the maximal transmit power $P_\text{max}$ is the summation of the occupation ratios of power for all $K$ sub-bands, which is given by
\begin{equation}
    \label{eq:power occupation}
    U_{P}(i)=\sum\limits_{j\in\{1,2,\dots,N\}\backslash\{i\}}\sum\limits_{k\in\{1,2,\dots,K\}} \dfrac{P(i,j,k)}{P_\text{max}}.
\end{equation}

The total resource occupation is the summation of power and sub-array usage ratios with equal weights. 
This definition ensures that RE is emphasized equally in terms of both power and sub-array utilization, enabling a balanced trade-off between the usage of different types of resources.
In particular, as discussed in~\cite{tang2014resource}, the equal weights can benefit the enhancement of overall RE without substantially sacrificing the improvement of RE pertaining to any resource.
Consequently, the overall resource usage for the UAV $i$ is defined as the mean value of its power and sub-array occupation ratios, and can be written as
\begin{equation}
    \label{eq: total consumption}
    U(i)=\dfrac{U_{P}(i)+U_{S}(i)}{2}.
\end{equation}

In addition to the consideration of resource usage minimization, to acquire good quality of services, the latency and packet loss are taken into account as well.
Since this work is dedicated to optimizing the resource allocation for the THz links within the UAV network, the latency of a data packet measures the time from the origin UAV (i.e., the UAV that collects the data) to the header UAV (i.e., the UAV that transmits this packet from the THz UAV network to the LEO satellite network).

For the THz link from the UAV $i$ to the UAV $j$, the latency of a data packet $\varpi$ is determined by the data rate of this link as well as the number of packets that have already been stored in the buffer prior to it. 
This packet arrives in the UAV $i$ in the time slot $\tau$ with a duration of $\Delta\tau$, if and only if this UAV receives $\varpi$ at the time instant $(\tau-1)\Delta\tau+\Delta\tau'$, where $\Delta\tau'$ falls within $\left[0,\Delta\tau\right)$.
Then, in the remaining time of this time slot, the maximal amount of data that can be transmitted through this link is $R_\tau(i,j)(\Delta\tau-\Delta\tau')$.
Hence, the packet $\varpi$ can be transmitted to the UAV $j$ in this time slot if $R_\tau(i,j)(\Delta\tau-\Delta\tau')\geq\omega(\varsigma_{\varpi,\tau}(i,j)+1)$, where $\varsigma_{\varpi,\tau}(i,j)$ measures the number of packets stored in the buffer before $\varpi$ in the time slot $\tau$, and $\omega$ denotes the size of a data packet.
Additionally, the distance $d_\tau(i,j)$ and the rate $R_\tau(i,j)$ can approximate to constants in a small $\Delta \tau$, as discussed in~\cite{liu2022deep}.
Consequently, the delay for this packet is obtained as
\begin{equation}
    \label{eq: latency if transmit}
    T_\varpi(i,j)=\dfrac{d_\tau(i,j)}{c}+\dfrac{\omega(\varsigma_{\varpi,\tau}(i,j)+1)}{R_\tau(i,j)}.
\end{equation}
On the contrary, if $R_\tau(i,j)(\Delta\tau-\Delta\tau')<\omega(\varsigma_{\varpi,\tau}(i,j)+1)$, the transmission of the packet $\varpi$ needs to further wait for $\varkappa$ time slots, where $R_{\tau+\varkappa}(i,j)\Delta\tau\geq\omega(\varsigma_{\varpi,\tau+\varkappa}(i,j)+1)$,
while $R_{\tau'}(i,j)(\Delta\tau-\Delta\tau')<\omega(\varsigma_{\varpi,\tau'}(i,j)+1),\forall\tau'\in\{\tau+1,\tau+2,\dots,\tau+\varkappa-1\}$.
In this case, the latency is given by
\begin{equation}
    \label{eq: latency if not transmit}
    T_\varpi(i,j)=\dfrac{d_{\tau+\varkappa}(i,j)}{c}+\dfrac{\omega(\varsigma_{\varpi,\tau+\varkappa}(i,j)+1)}{R_{\tau+\varkappa}(i,j)}+\varkappa\Delta\tau-\Delta\tau'.
\end{equation}
If the packet $\varpi$ from the origin UAV is not lost and successfully received by the header UAV, the overall delay on its entire transmission path $L_\varpi$ in the THz UAV network is calculated as
\begin{equation}
    \label{eq: total latency}
    T_\varpi=\sum\limits_{(i,j)\in L_\varpi} T_\varpi(i,j).
\end{equation}

Furthermore, the number of lost packets $l_\tau(i)$ for the UAV $i$ is determined by the data rate $R_\tau(i,j)$, the numbers of packets already in the buffer of this UAV $\varsigma'_\tau(i)$, as well as the number of incoming packets $\varrho_\tau(i,j)$.
In particular, the data rate and the storage capacity of the UAV $i$ can handle at most $\Omega+\left\lfloor\frac{R_\tau(i,j)\Delta \tau}{\omega}\right\rfloor$ packets at the time slot $\tau$ without packet loss, where $\lfloor\cdot\rfloor$ stands for the floor function.
In addition, the number of packets that need to be transmitted or stored is the summation of $\varsigma'_\tau(i)$ and $\varrho_\tau(i,j)$.
Therefore, if $\varrho_\tau(i,j)+\varsigma'_\tau(i)>\Omega+\left\lfloor\frac{R_\tau(i,j)\Delta \tau}{\omega}\right\rfloor$, the number of the lost packets for the UAV $i$ is expressed as
~\cite{hu2023deep}
\begin{equation}
    \label{eq: single UAV packet loss}
    l_\tau(i)=\varrho_\tau(i,j)+\varsigma'_\tau(i)-\Omega-\left\lfloor\frac{R_\tau(i,j)\Delta \tau}{\omega}\right\rfloor.
\end{equation}
In contrast, if $\varrho_\tau(i,j)+\varsigma'_\tau(i)\leq\Omega+\left\lfloor\frac{R_\tau(i,j)\Delta \tau}{\omega}\right\rfloor$, the data rate and the storage of packets can handle the summation of the data amounts of $\varsigma'_\tau(i)$ as well as $\varrho_\tau(i,j)$ and, thus, no packet loss occurs for the UAV $i$ in time slot $\tau$.
The total number of lost packets in time slot $\tau$ in the entire UAV network is given by
\begin{equation}
    \label{eq: total UAV packet loss}
    l_\tau=\sum\limits_{i\in\{1,2,\dots,N\}} l_\tau(i).
\end{equation}
 
For the joint power and sub-array allocation problem in the THz UAV network, the goal is to maximize RE in terms of the expected usage of power and sub-arrays averaged over all UAVs for a long-term period (i.e., including multiple time slots $\tau$ that commence from any time instant $t$).
As aforementioned, the latency and packet loss are considered as well, to ensure the quality of services.
Consequently, we model the objective function as  
\begin{subequations}
        \label{eq:total objective}
        \begin{align}
        \label{eq: object}
        &\mathop{\arg\min}\limits_{\pi_p\left({\boldsymbol{P}_\tau, \boldsymbol{S}_\tau}\right)}
        \sum\limits_{\tau=t}^\infty \kappa^{\tau-t} \mathbb{E}_{\pi_p}\left[\dfrac{\sum_{i=1}^N U_{\tau}(i)}{N}\right],\\
        \label{eq: single power constraint}
        \textrm{s.t.}\quad 
        & 0\leq P_\tau(i,j,k)\leq P_\text{max}, \quad\forall \tau,i,j,k,\\
        \label{eq: total power constraint}
        & 0\leq U_{P,\tau} (i)\leq1, \quad\forall \tau, i,\\
        \label{eq: single transmitting subarray constraint}
        & 0\leq S_{\text{Tx},\tau}(i,j)\leq S_\text{max}, \quad\forall \tau,i,j,\\
        \label{eq: single receiving subarray constraint}
        & 0\leq S_{\text{Rx},\tau}(i,j)\leq S_\text{max}, \quad\forall \tau,i,j,\\
        \label{eq: total subarray constraint}
        & 0\leq U_{S,\tau} (i)\leq1, \quad\forall \tau, i,\\
        \label{eq: delay constraint}
        & T_\tau\leq T_\text{max}, \quad\forall \tau,\\
        \label{eq: packet loss constraint}
        & l_\tau\leq l_\text{max}, \quad\forall \tau.
        \end{align}
\end{subequations}
In~\eqref{eq: object}, $\mathbb{E}[\cdot]$ represents the expected value.
$\kappa\in[0,1]$ is the attenuation coefficient for future resource occupations.
More concretely, a small $\kappa$ reflects a high weighted contribution of the current resource usage, while a large $\kappa$ underscores the future resource utilization.
In~\eqref{eq: single power constraint} and~\eqref{eq: total power constraint}, for every UAV in each time slot, the allocated transmit power for each band of each THz link, as well as the total utilized power, cannot exceed the maximal power provided by this UAV.
Similarly, the numbers of transmitting, receiving, and overall assigned sub-arrays should be less than or equal to the maximal sub-array count of each UAV in each time slot, formulated as~\eqref{eq: single transmitting subarray constraint},~\eqref{eq: single receiving subarray constraint}, and~\eqref{eq: total subarray constraint}.
In~\eqref{eq: delay constraint}, the latency in every time slot $T_\tau$ is required to remain within the tolerable limitation.
Specifically, $T_\tau$ is the latency averaged over all packets that successfully arrive at the header UAV in the time slot $\tau$.
Furthermore, the total number of lost packets in every time slot cannot be greater than the limitation, as shown in~\eqref{eq: packet loss constraint}.

As discussed in Sec.~\ref{sec: intro}, unlike terrestrial backhaul networks with fixed base stations as well as satellite networks with predictable satellite trajectories, the unpredictable variational position of each UAV results in frequent reconfigurations of links and routing in the THz UAV network.
This attribute incurs the highly dynamic topology of the entire UAV network, making the above long-term problem much more challenging.
Moreover, the highly intricate relationship between long-term resource allocation strategy and the factors in the objective function further imposes complications on this problem.
For example, the latency $l_\tau$ in each time slot $\tau$ depends on the data rates of all THz links, the dynamic routing paths, the transmission orders of a large number of packets, as well as the buffer storage status and the amount of incoming data for every UAV, in multiple time slots before $\tau$. 
Additionally, as aforementioned, the unpredictable stochastic data traffic demands of UAVs further nullify the feasibility of traditional non-learning algorithms.
Hence, the DRL algorithm with a powerful learning ability, which is tailored to the dynamic THz UAV network, is motivated to effectively solve the above issues. 
\section{Deep Reinforcement Learning Framework}
\label{sec: DRL description}

Aiming at addressing the NP-hard long-term joint power and sub-array allocation problem in~\eqref{eq:total objective} in the THz UAV network, we design a tailored DRL framework.
More specifically, the DRL learning agent works as follows.
In any time slot $\tau$, the DRL learning agent first observes the state $s_\tau$ (related to all UAVs and THz links) from the wireless environment, i.e., the entire THz UAV network.
By processing the perceived observation, the agent provides the power and sub-array allocation action $\Xi_\tau$ for each UAV, which supports the THz transmissions for the traffic demands.
After executing the resource allocation action, the agent can procure the reward $r_\tau$, which is determined by key factors in the objective function in~\eqref{eq:total objective} (namely, resource usage, latency, and number of lost packets).
Due to the changes in the environment in the next time slot, the observed state $s_\tau$ is updated to $s_{\tau+1}$.
Consequently, the action and reward are correspondingly updated to $\Xi_{\tau+1}$ and $r_{\tau+1}$, respectively.

Based on the experiences of state, action, and reward, the DRL agent is trained to optimize the resource allocation policy to pursue the highest long-term accumulated rewards.
For ease of the collection of latency for the packets that arrive at the header UAV, the agent is installed at the header UAV.
In this section, we elaborate the important elements of the proposed DRL framework, i.e., \textit{state}, \textit{action}, and \textit{reward}.

\subsection{State}
\label{sec: state}

To enable adaptive resource allocation decision-making on the basis of the observed state, the DRL agent in the header UAV monitors the following critical parameters about all links and UAVs in each time slot.

\begin{itemize}
    \item Each UAV needs to meet its traffic demand, including the data packets of the missions executed by this UAV, the data packets it relays, as well as the data packets stored in its buffer. 
    Although the stochastic data volumes are unpredictable and uncertain, their mean values can be acquired via statistical analysis.
    Given these mean values, together with the topology (provided by the resource-efficient routing that is discussed in Sec.~\ref{sec: problem formulation}) and the occupation ratio of the buffer, the expected number of packets to be transmitted by each UAV can be attained and provided to the agent.
    
    \item The data rate of each THz link determines the quality of services of the transmitted data.
    As shown in~\eqref{eq:rate}, the rate relies on the SINRs of the channels for all sub-bands.
    In addition, the SINRs are directly affected by the power and sub-array allocation action.
    Hence, the DRL agent collects the SINRs of THz links as well.
    Using THz channel estimation techniques, the channel state information can be obtained~\cite{chen2021hybrid}.
    
    \item Due to the time-varying UAV positions and network topology, the location information of each UAV and the transmission distance of each THz link are crucial for the data packet delivery as well.
    In this work, we assume that the UAVs fly at the same altitude and, thus, the horizontal position information is required to identify the location of each UAV.
    Additionally, each UAV estimates the distance from itself to the next hop.
    The location and distance information of the entire THz UAV network are gathered by the header UAV (i.e., the agent).
    Moreover, each UAV provides an indicator to the agent, to indicate whether it is the destination of all data packets in the THz UAV network, i.e., the header UAV.
\end{itemize}

Hence, the observed state in the time slot $\tau$ with the consideration of all the above factors is expressed as $s_{\tau}=\left\{s_{\tau}(i)|i\in\{1,2,\dots,N\}\right\}$~\cite{RLbook}.
In particular, the entry $s_{\tau}(i)=\left\{\bar\mu_\tau(i), \iota_\tau(i),\boldsymbol{\gamma}_\tau(i,j), d_\tau(i,j),x_\tau(i),y_\tau(i), \psi(i)\right\}$, where $\bar\mu_\tau(i)$ is the expected number of incoming packets that need to be sent by the UAV $i$ in the time slot $\tau$.
$\iota_\tau(i)$ stands for the occupation ratio of the buffer of the UAV $i$.
$\boldsymbol{\gamma}_\tau(i,j)$ symbolizes the SINR information of the THz link from the UAV $i$ to the UAV $j$ (i.e., the next hop of the UAV $i$) with the propagation distance $d_\tau(i,j)$.
$x_\tau(i)$ and $y_\tau(i)$ measure the horizontal location of the UAV $i$.
$\psi(i)$ is 1 or 0, indicating whether the UAV $i$ is the header UAV or not.

\subsection{Action}
\label{sec: action}

In line with the joint power and sub-array allocation problem in~\eqref{eq:total objective}, the allocation strategies of both power and sub-arrays in each UAV comprise the action.
The power and sub-array resources of each UAV are allocated to the THz links between this UAV and the UAVs connected to it.
In particular, the transmitting sub-arrays and power are allocated to support the transmission from this UAV to its next hop.
On the contrary, the receiving sub-arrays are assigned to enable the UAV to receive packets from previous hops.
The power and sub-array resource allocation actions are elaborated as follows.
\begin{itemize}
    \item The power allocation action of the UAV $i$ in a time slot $\tau$ consists of the assigned transmit power ratios for all sub-bands in the THz link from the UAV $i$ to its next hop.
    In particular, the allocation action is expressed as $\boldsymbol{P}_\tau(i)=\left\{\tilde P_\tau(i,k)|k\in\{1,2,\dots,K\}\right\}$, where $\tilde P(i,k)$ represents the ratio of occupied power for the $k^{th}$ sub-band, $\sum_{k\in\{1,2,\dots,K\}}\tilde P_\tau(i,k)\leq1$, and $\tilde P_\tau(i,k)\geq0,\forall k\in\{1,2,\dots,K\}$.
    The values of allocated power are the products of the assigned power ratios and the value of maximal power provided by the UAV.
    \item The sub-array allocation action for the UAV $i$ contains the ratios of allocated transmitting and receiving sub-arrays, which is given by $\boldsymbol{S}_\tau(i)=\left\{\tilde S_{\text{Tx},\tau}(i),\tilde S_{\text{Rx},\tau}(i)\right\}$, where $\left.\tilde S_{\text{Tx},\tau}(i)+\tilde S_{\text{Rx},\tau}(i)\leq1\right.$, and $\left.\tilde S_{\text{Tx},\tau}(i),\tilde S_{\text{Rx},\tau}(i)\geq 0\right.$.
    If the UAV $i$ relays signals for multiple UAVs, the allocated receiving sub-array ratio is evenly distributed to the corresponding THz links, as mentioned in Sec.~\ref{sec: problem formulation}.
    In contrast, if it does not serve as a relay, no sub-array is assigned for reception.
    Furthermore, since at least one transmitting sub-array and one receiving sub-array are needed to support a THz link, one sub-array is pre-assigned to each link for transmission or reception.
    As a result, the numbers of used transmitting and receiving sub-arrays, which are integers, are obtained by rounding down the products of the remaining number of sub-arrays and the allocated ratios.
\end{itemize}

As a result, the action in the time slot $\tau$ is defined as $\Xi_\tau=\left\{\boldsymbol{P}_\tau,\boldsymbol{S}_\tau\right\}$, where $\boldsymbol{P}_\tau=\left\{\boldsymbol{P}_\tau(i)|i\in\{1,2,\dots,N\}\right\}$, and $\boldsymbol{S}_\tau=\left\{\boldsymbol{S}_\tau(i)|i\in\{1,2,\dots,N\}\right\}$.
Benefiting from the tailored definition of the action, the power and sub-array constraints limiting the values of allocated resources in~\eqref{eq: single power constraint}, \eqref{eq: total power constraint}, \eqref{eq: single transmitting subarray constraint}, \eqref{eq: single receiving subarray constraint}, and~\eqref{eq: total subarray constraint} are satisfied.

\subsection{Reward}
\label{sec: reward}

In the context of DRL, the reward guides the learning process of the agent, to provide actions with higher rewards under the observed states.
According to the resource-efficient allocation problem in~\eqref{eq:total objective}, to encourage the DRL agent to generate the desired allocation policy, the reward needs to be designed to expedite resource usage minimization. 
Hence, in any time slot $\tau$, the instant reward is related to the additive inverse of the amount of current resource usage.
Moreover, apart from the power and sub-array constraints in~\eqref{eq: single power constraint}, \eqref{eq: total power constraint}, \eqref{eq: single transmitting subarray constraint}, \eqref{eq: single receiving subarray constraint}, and~\eqref{eq: total subarray constraint} that have been met via the designs of actions in Sec.~\ref{sec: action}, the remaining latency and packet loss constraints in~\eqref{eq: delay constraint} and~\eqref{eq: packet loss constraint} should be considered as well.
Thus, the reward consists of the incentive for resource occupation reduction, together with penalties for latency and packet loss, which is expressed as
\begin{equation}
    r_\tau=-\left(
    \chi_1 \dfrac{\sum_{i=1}^N U_{\tau}(i)}{N} + 
    \chi_2 T_\tau +
    \chi_3 l_\tau
    \right)
    ,
\label{eq:reward with panelty}
\end{equation}
where the scaling coefficient $\chi_1$ facilitates the DRL convergence during training by regulating the range of the reward.
$\chi_2$ and $\chi_3$ refer to the coefficients governing the penalties related to latency and packet loss, respectively.
To preclude catastrophic deterioration in the quality of services throughout the entire training process, large latency and severe packet loss should be prevented, benefiting the practical implementation of the DRL framework.
Nevertheless, state-of-the-art constraint reinforcement learning methods require a large number of episodes to mitigate actions that violate constraints~\cite{murti2022constrained,marchesini2022exploring,xu2021crpo,wu2020dynamic}. 
Even after convergence, the simulations of these approaches reveal that they cannot guarantee the elimination of constraint violations as well. 
Therefore, rather than employing constrained DRL techniques, $\chi_2$ and $\chi_3$ are set as sufficiently large values, encouraging the DRL to rigorously avoid the generation of undesired actions that result in severe latency and packet loss throughout training. 

Consequently, the long-term resource-efficient joint power and sub-array allocation problem in~\eqref{eq:total objective} can be equivalently reformulated as a long-term reward maximization problem within the DRL framework, which is given by
\begin{equation}
    \centering
    \label{eq:DRL problem}
    \begin{aligned}
    \mathop{\arg\max}\limits_{{\pi_p}\left(\Xi_\tau\right)}
\sum\limits_{\tau=t}^\infty\kappa^{\tau-t} \mathbb{E}_{\pi_p}\left[
r_\tau
\right].
     \end{aligned}
\end{equation}

\section{Graph Neural Network Aided Deep Reinforcement Learning for Resource Allocation with an Emphasis on Self-node Features}
\label{sec: GNN-DRL}

In this section, we propose a GLOVE algorithm to realize the long-term resource-efficient joint power and sub-array allocation in the THz UAV network.
Particularly, GLOVE maximizes the long-term reward in the context of the aforementioned DRL framework.
GLOVE takes charge of optimizing the resource allocation strategy for continuous power and sub-array ratios, by building on the deep deterministic policy gradient (DDPG) algorithm, which is specifically designed for continuous action space with the actor-critic structure~\cite{lillicrap2015continuous}.
In the GLOVE algorithm for the THz UAV network, the actor determines the joint power and sub-array allocation action based on the observed state in each time slot.
Then, the action is assessed by the critic via the Q value, which measures the highest potential accumulated future reward over a long-term period.
To learn from the graph information while quickly capturing the important self-node state when generating allocation action, we design a GNN-aided structure with an emphasis on self-node features.
The overall GLOVE learning framework, its key components (namely, actor and critic), as well as the GLOVE algorithm, are elaborated as follows.

\subsection{Overall Learning Framework}
\label{sec: overall learning framework}

\begin{figure}[t]
\centering
        \includegraphics[width=\linewidth]{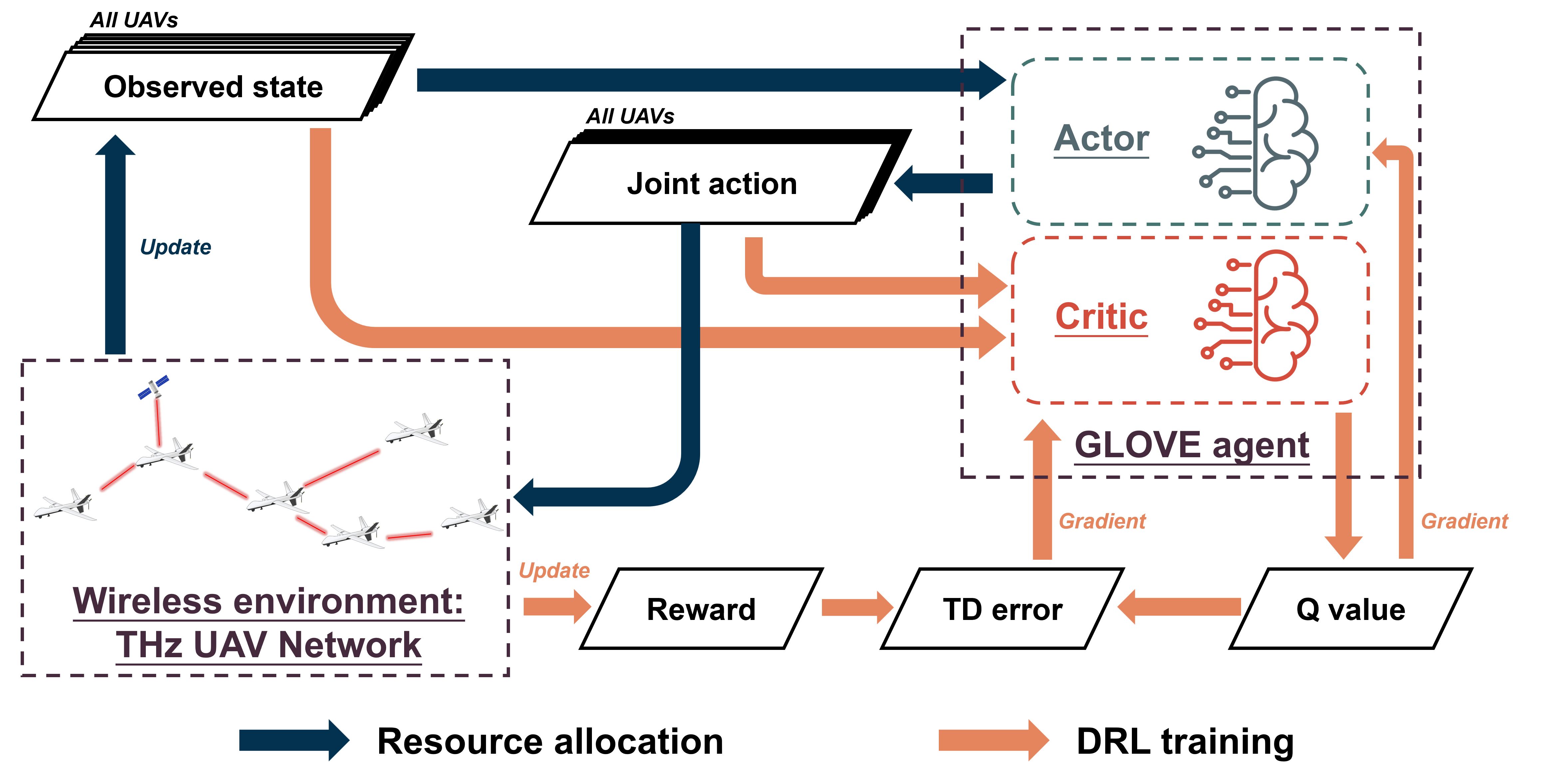} 
        \caption{{Allocation and learning process of GLOVE.}}
        \label{fig: GLOVE flow}
\end{figure}

As depicted in Fig.~\ref{fig: GLOVE flow}, GLOVE provides the joint power and resource allocation actions in the THz UAV network (i.e., environment).
To optimize the allocation policy, GLOVE cooperatively trains its actor and critic via the interplay between the agent and wireless environment.
In particular, the actor first collects the important parameters of each UAV and each link to construct the observed state (in the form mentioned in Sec.~\ref{sec: state}).
Then, the actor generates joint power and sub-array resource allocation action (in the form mentioned in Sec.~\ref{sec: action}) for each UAV, which is customized for the current observed state.
To evaluate the performance of the action in terms of the long-term reward under the current state, the critic processes the state-action pair and generates the Q value.

Therefore, the learning process of GLOVE relies on the exchange of critical parameters between the header UAV and others.
Under the float32 data type, the data size of each sample of the DRL experience (including state, action, and reward) for every UAV in each time slot for DRL learning is 76 bytes. 
Given the relatively small data size, the narrow-band feedback channel~\cite{lu2009simple} can efficiently handle the transmission of experiences.

Furthermore, for the purpose of high adaptability to the dynamic topology, on-policy training is adopted in GLOVE.
Under the on-policy training, the agent is trained with the current DRL experience.
Therefore, the policy can focus on and rapidly adapt to the current topology and environment, without the time-consuming collection of historical experiences that might be outdated in the highly dynamic THz UAV network.
Moreover, the on-policy training is especially suitable for THz UAV networks in hostile and harsh regions, where it is impractical for the agent to collect abundant samples of experiences for training. 
With the on-policy training, the GLOVE agent generates the joint resource allocation action with the current observed state of the THz UAV networks and allocation policy.
Then, the instant reward related to RE, latency, and packet loss is perceived by the agent.
In the next time slot, the state is updated.
The state, action, instant reward, as well as the next state, are fed into the GLOVE agent to optimize the allocation policy.
On the basis of the updated state and policy, the action and reward are accordingly updated.
With the updated experience in each time slot that is generated via the interaction between the GLOVE agent and the wireless environment, the on-policy training process repeats.

\subsection{GNN-aided Actor and Critic Networks with an Emphasis on Self-node Features}
\label{sec: actor & critic}

\begin{figure*}[t]
\centering
        \includegraphics[width=\linewidth]{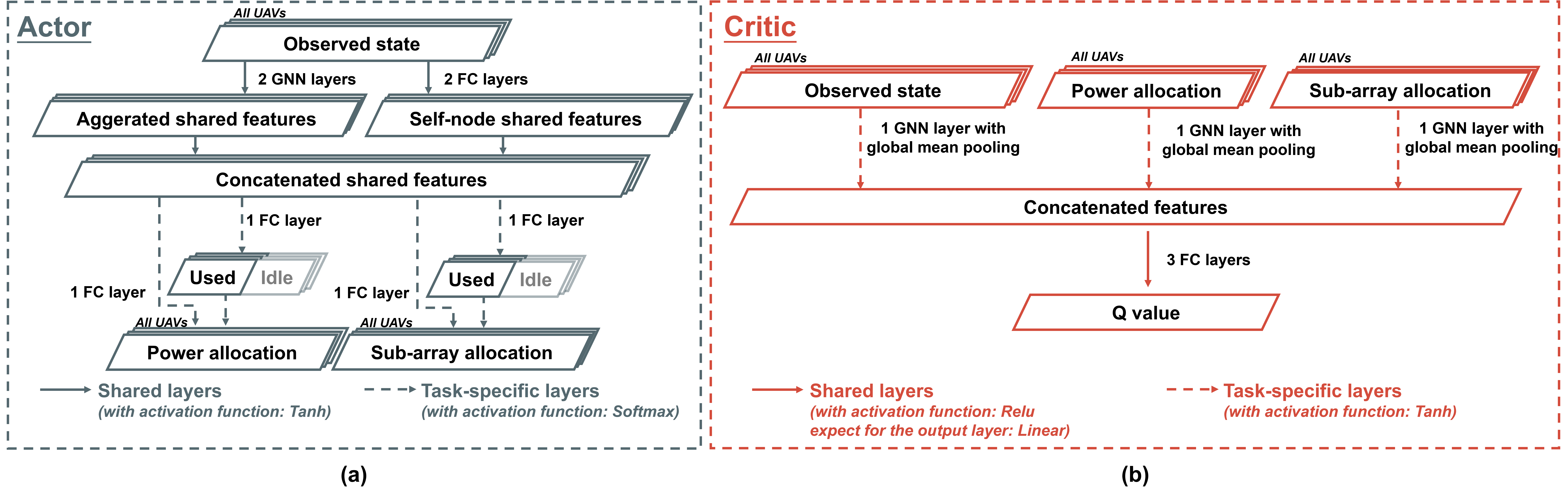} 
        \caption{GLOVE agent architectures (a) of the actor network, and (b) of the critic network.}
        \label{fig: GLOVE actor and critic}
\end{figure*}

The actor and critic comprise the GLOVE agent in the header UAV.
As shown in Fig.~\ref{fig: GLOVE actor and critic}(a), the actor perceives the observed state of the THz UAV network.
By processing the state with the parameters in the actor network, the actor provides the resource allocation results for long-term RE maximization (or equivalently, reward maximization).
To expedite the cooperative training of the tasks for power allocation and sub-array allocation, the GLOVE actor adopts the multi-task structure with shared and task-specific layers.
On one hand, the shared layers capture shared knowledge across both tasks.
On the other hand, the task-specific layers utilize the shared features to generate the respective outputs for the power and sub-array allocation actions simultaneously.
Furthermore, the backpropagation process for the training of both the power and sub-array allocation tasks can concurrently assist the shared layers in extracting useful features from the observed state.

In line with the definitions of power and sub-array allocation actions in Sec.~\ref{sec: action}, Softmax activation functions are leveraged to output the assigned power and sub-array ratios.
In particular, the Softmax function generates nonnegative values, whose summation is 1.
To adjust the amount of overall occupied power and sub-arrays, a Softmax function is used to determine the utilized and unused proportions for each type of resource.
Then, the utilized ratio is multiplied by the output of another Softmax function, allocating the utilized power/sub-arrays to different sub-bands/links.
Therefore, the values of allocated resources are nonnegative and cannot exceed the limitations, ensuring the power and sub-array constraints in~\eqref{eq: single power constraint}, \eqref{eq: total power constraint}, \eqref{eq: single transmitting subarray constraint}, \eqref{eq: single receiving subarray constraint}, and~\eqref{eq: total subarray constraint}.

GNN is adopted in the GLOVE actor as well, since the topology of the THz UAV network is important for the THz transmissions from the origin UAVs to the header UAV.
More specifically, a classic GNN structure, namely, the graph convolutional network (GCN), is deployed to exploit the graph information in each time slot.
GCN layers can allow each UAV node to aggregate and refine features from adjacent nodes based on the connectivity information of the THz UAV network.
In contrast to multi-agent structures that assign a unique agent to each UAV, as well as the approaches that directly concatenate features of all UAVs, GCN layers enable each UAV node to share the same parameters for learning.
As a result, the number of trainable parameters of GCN layers remains invariant regardless of the graph size.
Hence, the memory and computational overheads of GLOVE can be alleviated, ensuring high scalability in the multi-UAV scenario.

For each GCN layer with the parameters $\mathcal{W}$, the activation function $\sigma_a(\cdot)$, as well as the input features of all UAV nodes $F$ (whose $i^\text{th}$ row is features of the UAV $i$), the output features can be expressed as 
\begin{equation}
\label{eq: GCN message passing}
F'= \sigma_a\left( \tilde{D}^{-\frac{1}{2}} \tilde{A} \tilde{D}^{-\frac{1}{2}} F \mathcal{W} \right),
\end{equation}
where $\tilde{A}$ stands for the adjacency matrix with self-loops that aggregate the features of each UAV and its neighbors.
More concretely, $\tilde{A}$ is defined as the sum of the adjacency matrix $A$ and the identity matrix $I$.
To capture the information from both the upstream and downstream nodes, $A_{i,j}=1$ if the UAV $i$ is the parent or child node of the UAV $j$ under the current routing result. 
Otherwise, $A_{i,j}=0$.
$\tilde{D}$ denotes the degree matrix of $\tilde{A}$, which is a diagonal matrix showing the number of direct connections (i.e., THz links) each node has within the graph.
Its $i^\text{th}$ entry is expressed as $\tilde{D}_{ii}=\sum_j \tilde{A}_{i,j}$.

Moreover, the transmission rate for the link between each UAV node and its next hop is predominantly affected by the self-node state (i.e., the state of the UAV node itself), which includes the crucial parameters for the channel capacity of this link, e.g., channel state information and propagation distance, among others.
Consequently, the self-node state is directly related to the resource-efficient allocation decision (including the transmitting sub-array and transmit power assignments) with good quality of services for this node.
Nevertheless, GNN might need many iterations to learn the direct influence of the self-node under the aggregated features. 
Instead, we design a structure in the actor to emphasize the self-node features in the entire DRL training process.
Particularly, in addition to the GCN layers to explore the information of the graph, fully-connected (FC) layers are adopted to extract features of each UAV without feature aggregation.
The GLOVE actor concatenates the aggregated and self-node features with the same dimensionality, which are obtained via the GCN and FC layers, respectively.
Then, the concatenated features are processed by the task-specific layers to generate the joint allocation action.
As a result, the GLOVE actor can benefit from not only structural relationships among nodes but also the critical local state when it allocates resources for each UAV node during the whole training period.

Apart from the GLOVE actor that provides allocation action based on the observed state, the GLOVE critic evaluates the action under the current state by inputting the state-action pair.
As depicted in Fig.~\ref{fig: GLOVE actor and critic}(b), owing to the multi-task structure of the actor, the critic deploys task-specific layers to extract features from the state, power allocation action, as well as sub-array assignment action.
For ease of feature extraction under the current topology, GCN layers are employed in the GLOVE critic as well.
Then, these features are concatenated and processed via shared layers, to generate the Q value that measures the potential maximal future accumulated reward, which is expressed as
\begin{equation}
\label{eq:Q}
\begin{aligned}
    Q(s_{t}, \Xi_{t})=&\max\limits_{\left\{\pi_p(\Xi_\tau)|\tau>t\right\}}\mathbb{E}\Bigg[\sum\limits_{\tau=t}^{\infty}\kappa^{\tau-t} r_\tau|\left(s_{t},\Xi_{t}\right)\Bigg].
\end{aligned}
\end{equation}
A higher Q value suggests that under the current state, the action generated by the DRL actor can lead to a higher long-term reward.

\subsection{GLOVE Algorithm}
\label{sec: GLOVE algorithm}

\begin{algorithm}[t]
\caption{GLOVE Training.}
\KwIn{Network parameters for the GLOVE actor and critic: $\theta_{a}$ and $\theta_{c}$, respectively
}
\KwOut{Well-trained joint resource allocation policy $\pi_p$
}

Initialize the observed state $s_{t}$

Initialize the actor network parameters $\theta_{a}$ to allocate most of the power and sub-arrays

    \For{$\tau=t,t+1,\dots$}
    {
        Generate the action $\Xi_{\tau}$ according to the current state $s_{\tau}$ and the actor network parameters $\theta_{a}$ with random action noises
        
        Calculate the reward $r_\tau$ for the state-action pair $\left(s_{\tau},\Xi_{\tau}\right)$

        Calculate $Q(s_{\tau},\Xi_{\tau}|\theta_{c})$ according to the critic network parameters $\theta_{c}$
        
        Update the state to $s'_{\tau+1}$ according to the action $\Xi_{\tau}$
        
        Calculate $\Xi'_{\tau+1}={\pi_p}(s'_{\tau+1}|\theta_{a})$

        Calculate $Q(s'_{\tau+1},\Xi'_{\tau+1}|\theta_{c})$ according to the critic network parameters $\theta_{c}$
        
        Calculate $
        y_\tau=r_\tau+Q(s'_{\tau+1},\Xi'_{\tau+1}|{\theta}_{c})
        $
        
        $\theta_{a}\leftarrow$ Adam gradient ascent $Q(s_{\tau},\Xi_{\tau}|\theta_{c})$ 
        
        $\theta_{c}$ $\leftarrow$ Adam gradient descent $
        [y_\tau-Q(s_{\tau},\Xi_{\tau}|\theta_{c})]^2$
    }

\end{algorithm}

The GLOVE algorithm contains the cooperative training of the actor and the critic via the interaction between the agent and the THz wireless environment.
On one hand, the actor is trained to generate the joint power and sub-array allocation action with the maximal Q value, expediting the long-term reward maximization, or equivalently, the long-term RE maximization.
Therefore, the goal of the training for the actor is to adjust its parameters $\theta_a$ via the gradient ascent of the Q value.
On the other hand, the critic is trained to provide the Q value that accurately assesses the action.
The optimal function of the Q value for any state-action pair, $Q^*$, satisfies the Bellman equation, which is given by
\begin{equation}
\label{eq:bellman function}
    Q^*(s_{\tau},\Xi_{\tau})=\mathbb{E}[r_\tau+\kappa \max\limits_{\Xi'_{\tau+1}} Q^*(s_{\tau+1},\Xi'_{\tau+1})|(s_{\tau},\Xi_{\tau})].
\end{equation}
Hence, the critic is trained to minimize the temporal difference (TD) error with the loss function that is given by
\begin{equation}
        \begin{aligned}
                \label{eq:loss critic}
            Loss(\theta_c)=&\mathbb{E}\Big[\Big(r_\tau+\kappa Q(s_{\tau+1},\Xi_{\tau+1}|{\theta_c}) -Q(s_{\tau},\Xi_{\tau}|\theta_c)\Big)^2\Big],
        \end{aligned}
        \end{equation}
where $\theta_c$ represents the parameters in the critic network.
The learning procedure of the GLOVE algorithm is detailed in Algorithm 1.

The overall computational complexity of the proposed algorithm is the summation of the complexities of all GCN and FC layers in the GLOVE actor and critic.
Concretely, with the input size $\tilde{F}$ and output size $\tilde{F'}$, the complexity of a GCN layer is $O((\tilde{E}+N)\tilde{F}\tilde{F'}+\tilde{E}\tilde{F})$~\cite{you2020l2}, which can be simplified as $O((\tilde{E}+N)\tilde{F}\tilde{F'})$, where 
$\tilde{E}$ THz links support the UAV network with $N$ UAV nodes.
In contrast, without the feature aggregation process, the complexity of the FC layer is $O(\tilde{F}\tilde{F'})$~\cite{hu2024multi}.

One step further, during the entire on-policy training process, the GLOVE algorithm needs to mitigate the risks of resource allocation actions with severe latency and packet loss that hinder efficient THz transmission.
Hence, the \textit{safe initialization} and \textit{safe exploration}~\cite{hu2023deep} mechanisms for the DDPG algorithms are adopted in GLOVE.
Under the safe initialization mechanism, the parameters of the actor network are configured to use most of the available resources initially, to safeguard against the high latency and a large number of lost packets during the early stage of learning.
Additionally, Gaussian noises are added to the DDPG action values to encourage explorations beyond the local optimum.
During the entire training process, the safe exploration mechanism can mitigate the calamitous explored actions with constraint violations on the fly.
Particularly, the noises for each type of resource in a UAV are forced to sum to zero, in case the explored amounts of allocated power or sub-array exceed the corresponding limitations. 
Furthermore, if the added noises result in negative allocated ratios for any link, they are discarded to alleviate the risks that some links fail to meet the traffic demands of UAVs.

\section{Simulation Results and Analysis}
\label{sec: result}

In this section, numerical results are presented to assess the performance of our proposed GLOVE algorithm in the dynamic THz UAV network on the fly.
Specifically, the performance metrics include key factors of the DRL reward in~\eqref{eq:reward with panelty}, namely, RE, latency, as well as the number of lost packets.
Moreover, accounting for the realistic deployment of the proposed algorithm, its running time and memory demand are evaluated as well.

\subsection{Simulation Parameters of THz UAV Network}
\label{sec: simulation parameters}

\begin{figure*}[t] 
\centering
        \includegraphics[width=\linewidth]{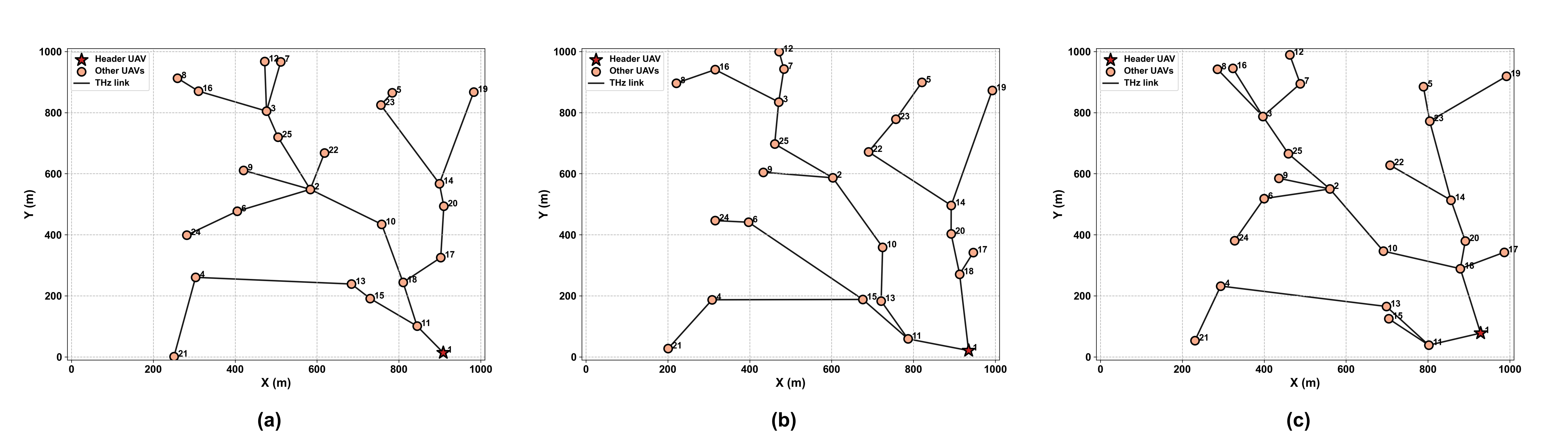} 
        \caption{THz UAV network topologies (a) at 0~s, (b) at 10~s, and (c) at 20~s.}
        \label{fig: UAV topology}
\end{figure*}

The simulations are implemented on Pytorch 1.12.1, which is installed on Python 3.8.0 with AMD Ryzen ThreadRipper 3990X CPU and Ubuntu 18.04. 
In our simulations, a THz UAV network in a $1000\times1000~\text{m}^2$ mission region is considered.
This network consists of $N=25$ UAVs with an altitude of 100~m~\cite{zeng2017energy} and a maximal speed of 100~m/s~\cite{fawaz2018uav}.
The horizontal positions of the UAVs are randomly initialized.
Among these UAVs, one UAV is randomly chosen as the header UAV to connect the THz UAV network with a LEO satellite.
By considering the unpredictable movements of terrestrial devices and surveilled targets (as mentioned in Sec.~\ref{sec: intro}), the UAVs follow a random motion pattern in the mission region.
Particularly, each UAV moves in a random direction and at a random speed that is lower than its maximal speed in each time slot~\cite{qin2024drl}.
As measured in our experiments, since the training time for GLOVE is 79~ms, the duration of each time slot is set as 0.1~s.
Each UAV can establish THz links with other UAVs within a maximal distance of $d_\text{max}=500$~m.
As demonstrated in Fig.~\ref{fig: UAV topology}, due to the frequently and significantly changing position of each UAV, the topology of the THz UAV network is highly dynamic under the resource-efficient routing scheme discussed in Sec.~\ref{sec: problem formulation}.

In this THz UAV network, the mean traffic demand for each UAV is 10~Gbps.
The size of each data packet is $\omega=2000$~bytes~\cite{xia2021multi}.
Moreover, each UAV deploys a buffer that can accommodate up to $\Omega=5\times10^5$ data packets.
To support the THz transmissions in the UAV network, $K=5$ nonoverlapped sub-bands with a bandwidth of $B=5$~GHz from 287.5~GHz to 312.5~GHz are utilized in each THz link.
Every UAV is equipped with $S_\text{max}=64$ planar sub-arrays.
Each sub-array contains $M_x\times M_y=4\times 4$ antennas.
The antenna gain for both the transmitting and receiving antennas is $G_\text{Tx}=G_\text{Rx}=5$~dBi.
Additionally, the maximal transmit power for every UAV is $P_\text{max}=30$~dBm.

\subsection{Performance Evaluation}
\label{sec: performance}

\begin{table}[tp]
\centering
\caption{Hyperparameters for GLOVE training.}
\begin{tabular}{@{}lc@{}}
\toprule
\textbf{Hyperparameter}                        & \textbf{Value}                              \\ \midrule
Training steps                                  & 1000                                  \\
Attenuating coefficient $\kappa$ & 0.5                                  \\
Variance of action exploration noises\qquad\quad~                      & $5\%\times$ allocated ratios\\
Learning rate for actor                                  &  $2\times10^{-5}$       \\
Learning rate for critic                               &  $10^{-2}$       \\
Scaling factor $\chi_1$                        & 10                                  \\
Penalty weight for latency $\chi_2$     & 5000                               \\
Penalty weight for packet loss $\chi_3$     &     $10^{-1}$                           \\
\bottomrule
\end{tabular}
\label{tb:hyper}
\end{table}

We implement the GLOVE algorithm with the hyperparameters that are summarized in Table~\ref{tb:hyper}.
These values are obtained by tuning, to ensure that the DRL training converges.
Then, we compare the performance of GLOVE with benchmark algorithms in terms of RE, latency, and packet loss on the fly, using these hyperparameters and the same UAV network configuration in each training step.
The benchmark schemes employ the on-policy DRL training strategy as well, to ensure their applicability for the long-term resource allocation problem in the highly dynamic THz UAV network.
In particular, the benchmark algorithms include the GNN-aided DRL methods with and without an emphasis on self-node features, as well as the multi-agent DRL approaches that cooperatively train the resource allocation policies of all UAVs without GNN, which are elaborated as follows.
\begin{itemize}
    \item GLOVE-actor-critic (GLOVE-AC) replaces the DDPG architecture with the structure of the actor-critic (AC) algorithm.
    Different from the DDPG-based algorithms that deterministically generate continuous values, AC is designed to select discrete actions in a stochastic manner.
    Specifically, the output layer of AC outputs the probability for the choice of each discrete action value. 
    Therefore, the power and sub-array allocation action spaces of GLOVE-AC are discretized.
    By considering the power and sub-array constraints in~\eqref{eq: single power constraint}--\eqref{eq: total subarray constraint}, the power and sub-array allocation actions are picked from $\left\{0, \frac{\bar{P}}{9},\frac{2\bar{P}}{9}, \dots,\bar{P}\right\}$ and $\left\{0, \frac{\bar{S}}{9},\frac{2\bar{S}}{9}, \dots,\bar{S}\right\}$, where $\bar{P}$ and $\bar{S}$ denote the values of power and sub-arrays under the uniform allocation, respectively.
    In addition, GLOVE-AC adopts the same GNN-aided structure with an emphasis on self-node features as GLOVE, except for the tailored output layer to select discrete actions.
    \item GNN-DDPG~\cite{deng2022resource} provides a deterministic policy to generate continuous allocated resource ratios by integrating GNN and the DDPG structure. 
    Different from GLOVE, GNN-DDPG lacks the FC layers that emphasize the self-node features of UAVs, while the architectures of other layers and the action spaces are the same as those of GLOVE.
    \item GNN-AC~\cite{dong2021intelligent} is a stochastic DRL structure that selects discrete resource allocation actions with the utilization of GNN layers.
    Without an emphasis on self-node features, GNN-AC has the same structure as GLOVE-AC, except for the exclusion of the FC layers extracting self-node features.
    Additionally, GNN-AC uses the same discrete action spaces as GLOVE-AC.
    \item Multi-agent DDPG (MADDPG)~\cite{gao2021game} deploys a unique DDPG agent in each UAV to determine the continuous power and sub-array ratio allocation without GNN.
    In the simulations, each MADDPG actor follows the same action spaces as well as the same architectures of shared and task-specific FC layers as the GLOVE actor.
    Since the reward in~\eqref{eq:reward with panelty} is related to the performance of the entire THz UAV network, one centralized critic is adopted to generate the Q values for the allocation actions of all agents.
    Moreover, the MADDPG critic replaces the GNN layers of the GLOVE critic with FC layers.
    \item Multi-agent AC (MAAC)~\cite{araf2022uav} exploits a unique AC actor to choose the resource allocation actions for each UAV, as well as a centralized critic to evaluate the actions of all UAVs.
    In particular, each MAAC actor employs the same FC layers as the GLOVE-AC actor, while discarding the GNN layers of the GLOVE-AC actor.
    Additionally, FC layers are leveraged in the MAAC critic to replace the GNN layers of the GLOVE-AC critic.
    Moreover, the MAAC critic uses the attention mechanism to cooperatively train all agents, which can learn the importance of the state-action pair of each UAV~\cite{iqbal2019actor}.
    The action spaces of MAAC are identical to those of GLOVE-AC.
\end{itemize}

\begin{figure}[t] 
\centering
        \includegraphics[width=\linewidth]{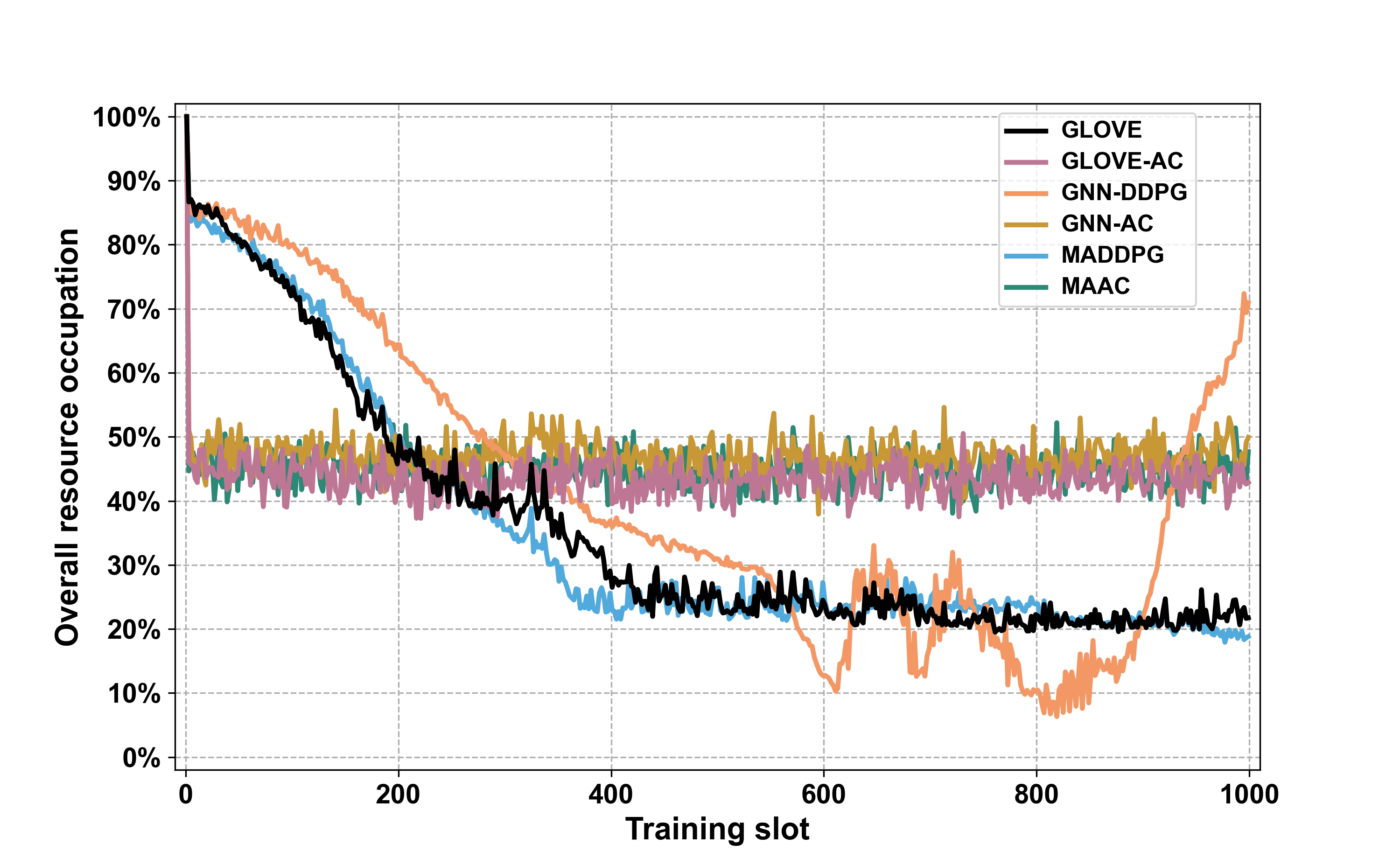} 
        \caption{Resource usage ratio comparison.}
        \label{fig: resource}
\end{figure}
\begin{figure}[t] 
\centering
        \includegraphics[width=\linewidth]{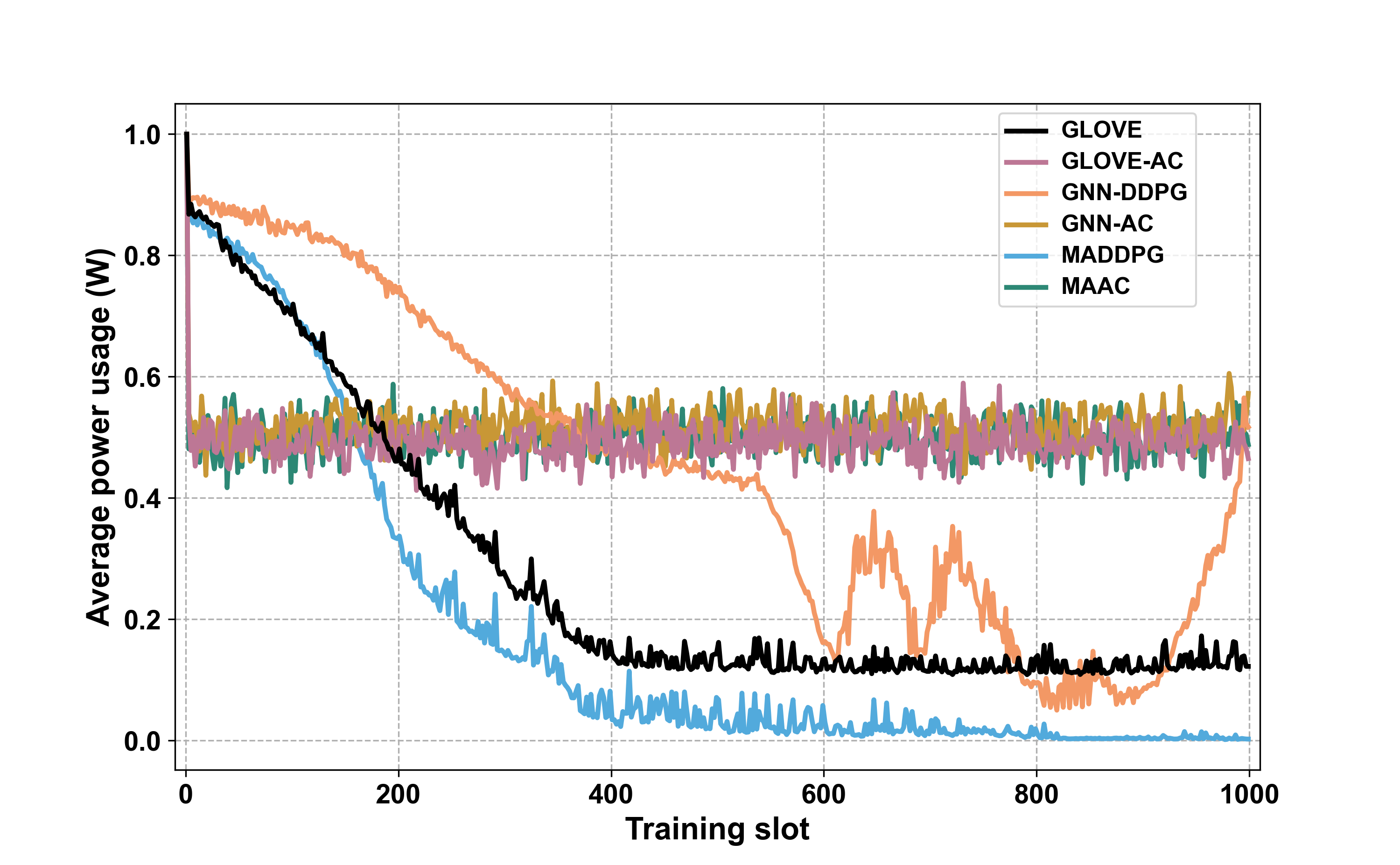} 
        \caption{Power usage comparison.}
        \label{fig: power}
\end{figure}
\begin{figure}[t] 
\centering
        \includegraphics[width=\linewidth]{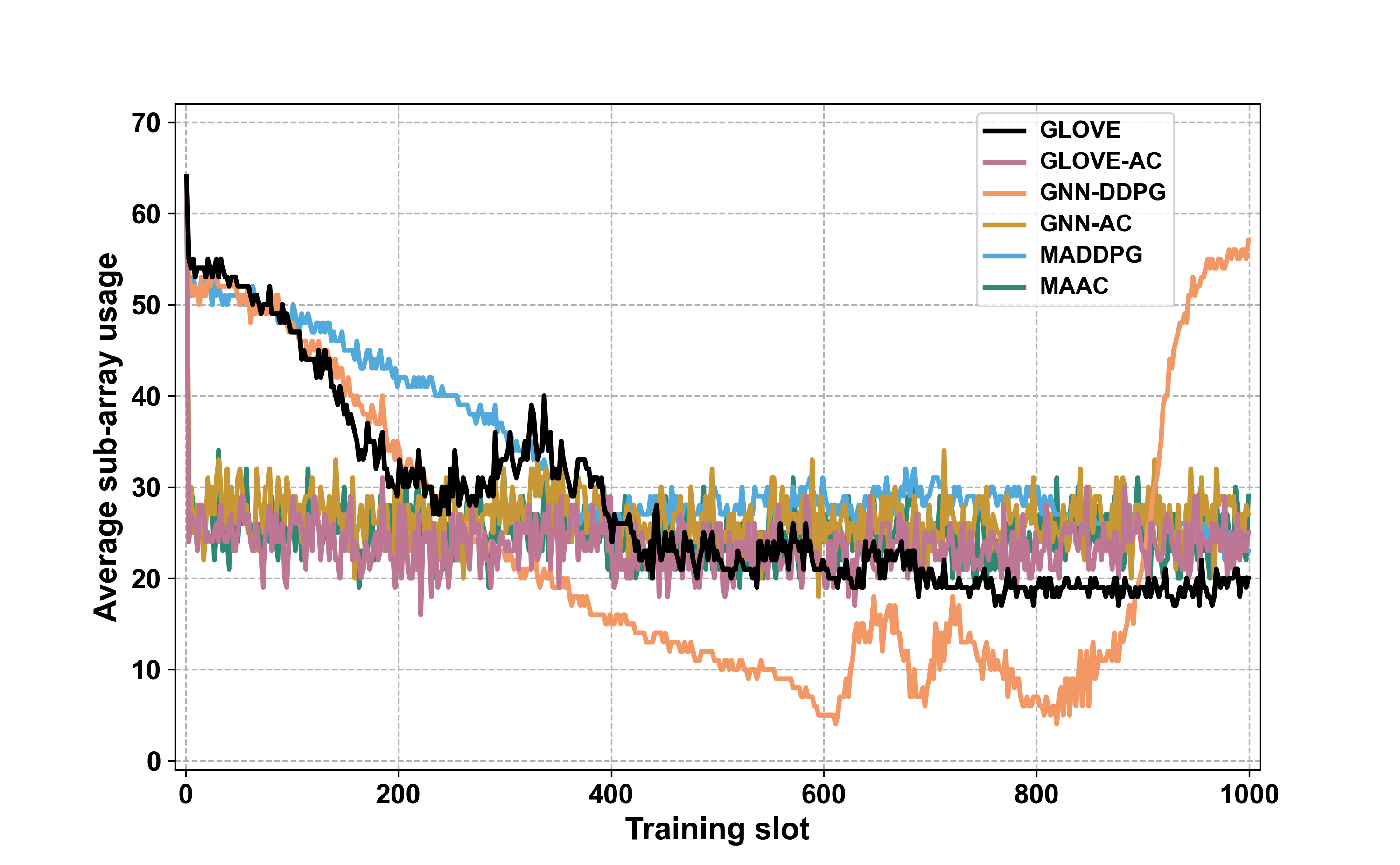} 
        \caption{Sub-array usage comparison.}
        \label{fig: array}
\end{figure}

As illustrated in Fig.~\ref{fig: resource}, the overall resource usage ratios of all AC-based algorithms fluctuate within the range of $[37\%, 55\%]$ in the entire training process, while the DDPG-based algorithms realize lower resource occupation during training.
Particularly, without the emphasis on self-node features, the resource usage of GNN-DDPG declines more slowly in the beginning, in comparison with GLOVE and MADDPG, which use FC layers to directly extract self-node features.
Then, the resource usage ratio of GNN-DDPG fluctuates within $[6\%, 33\%]$ from the 530th to the 870th training steps.
After the 870th training step, the ratio significantly increases and becomes the highest.
On the contrary, both GLOVE and MADDPG converge at the 20\% resource usage ratio.
Consequently, compared to GLOVE and MADDPG, AC-based algorithms use at least 69\% more resources, suggesting the superior RE of GLOVE and MADDPG.

The power and sub-array usage averaged over all UAVs are shown in Fig.~\ref{fig: power} and Fig.~\ref{fig: array}, respectively. 
As depicted, their trends resemble that of the overall resource usage.
The AC-based algorithms occupy more than 0.40~W power and 20 sub-arrays without achieving convergence.
In contrast, GLOVE converges at 0.12~W power and 19 sub-array usage.
Moreover, after MADDPG converges, it uses 0.10~W lower power and 7 more sub-arrays compared to GLOVE.
Although GNN-DDPG can achieve lower power and sub-array usage than GLOVE during training, it cannot converge and finally occupies the 0.57~W power and 57 sub-arrays, which are significantly higher than those of GLOVE.

\begin{figure}[t] 
\centering
        \includegraphics[width=\linewidth]{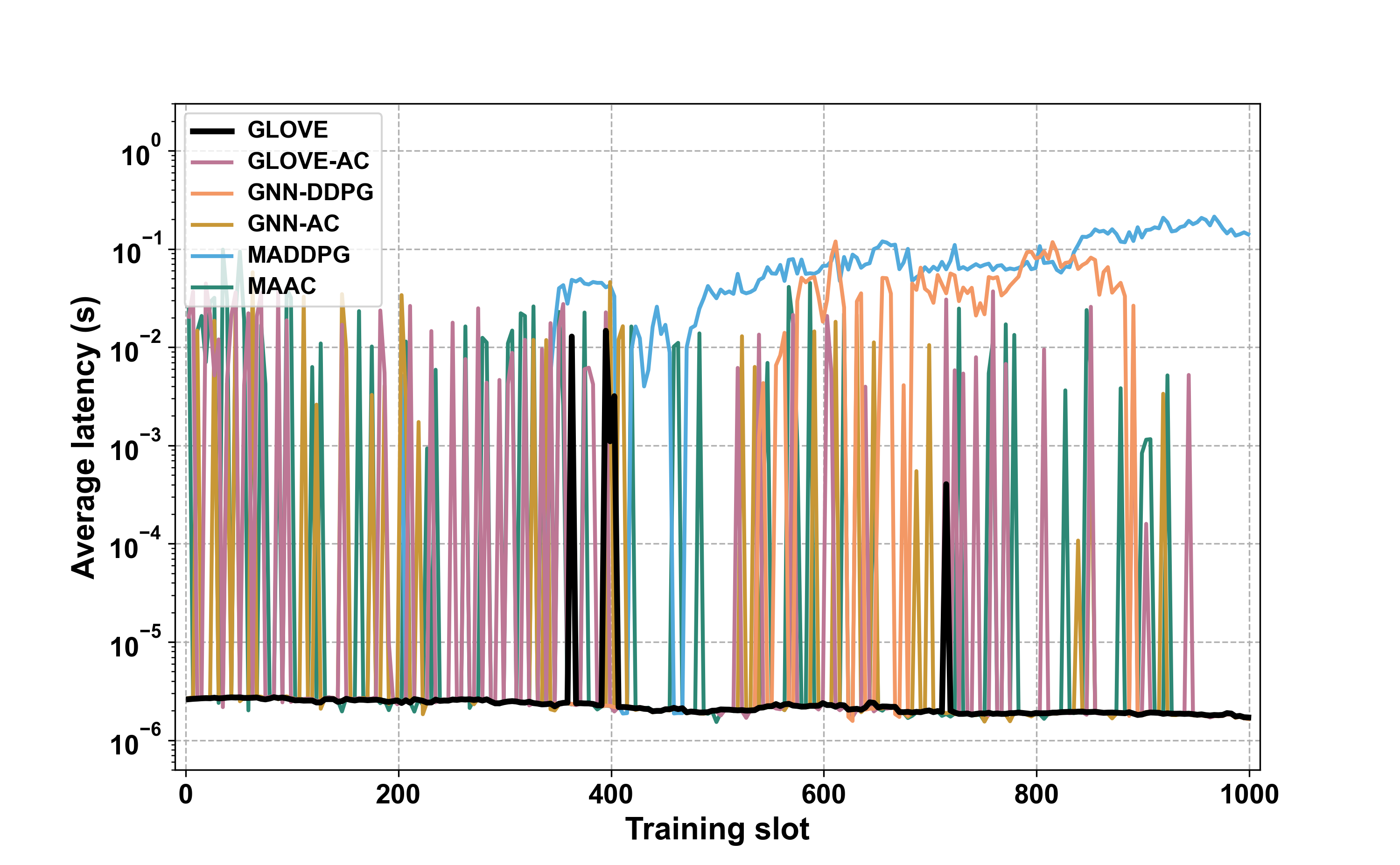} 
        \caption{Latency comparison.}
        \label{fig: latency}
\end{figure}

\begin{figure}[t] 
\centering
        \includegraphics[width=\linewidth]{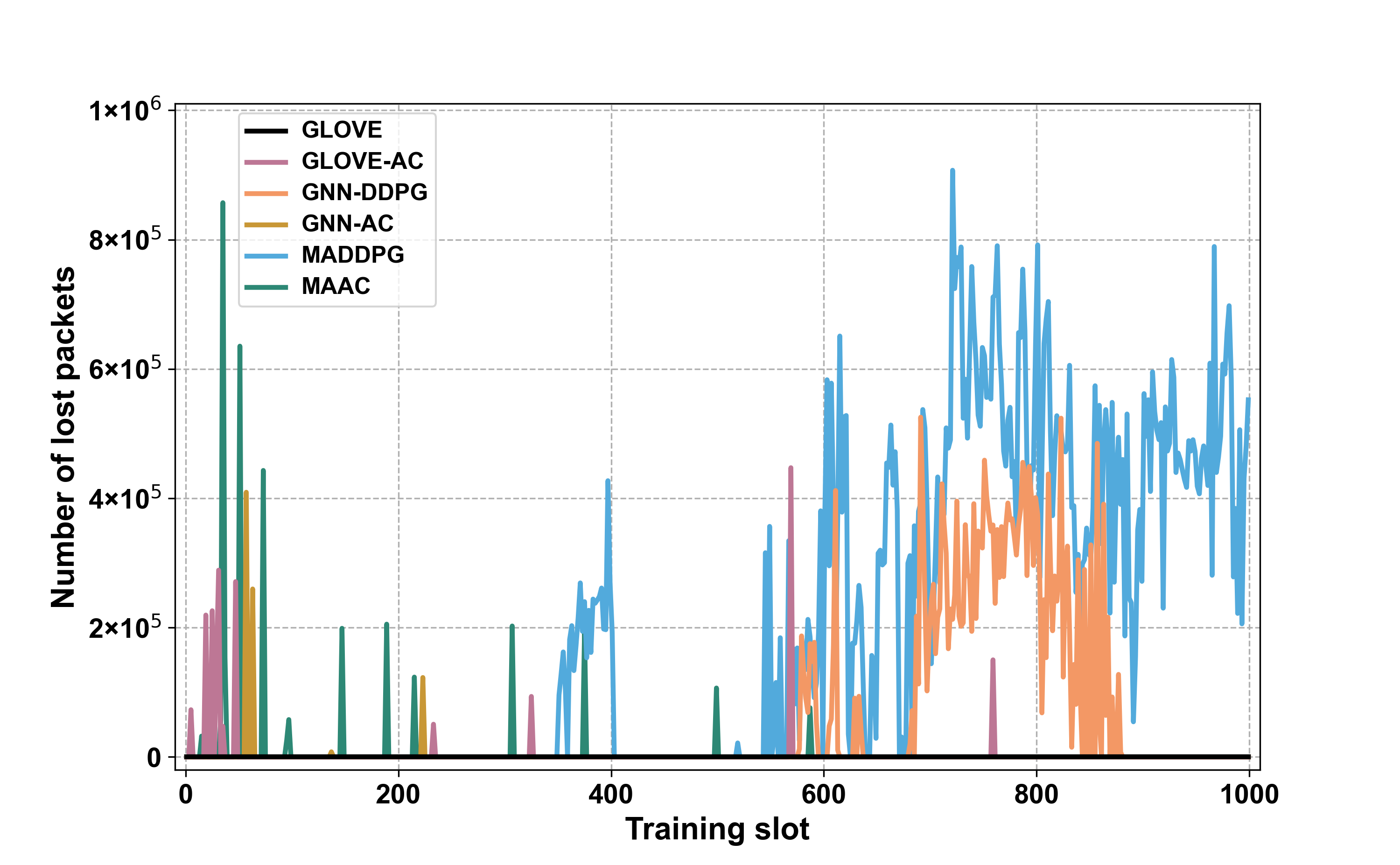} 
        \caption{Packet loss comparison.}
        \label{fig: packet loss}
\end{figure}

In addition to the resource usage, the latency (averaged over all data packets that successfully arrive at the header UAV in each time slot) and packet loss are important for the data delivery in the THz UAV network, which are demonstrated in Fig.~\ref{fig: latency} and Fig.~\ref{fig: packet loss}, respectively.
As the resource usage ratio of GLOVE decreases, ten-millisecond-level latency arises, primarily due to the delay caused by the data packets waiting in the buffer, which cannot be transmitted promptly.
After several occurrences of such latency, GLOVE converges and alleviates the delay incurred by data packets waiting in the buffer, by training with both the self-node features and the information extracted from the network topology.
The maximal latency of GLOVE is 15~ms.
In addition, GLOVE prevents packet loss throughout the whole training process.
On the contrary, all other algorithms frequently encounter latency ranging from 46~ms to hundreds of milliseconds and even a large number of lost packets (more than $4\times10^5$), leading to a catastrophe for transmissions in the THz UAV network. 
Among these algorithms, GNN-DDPG can learn to increase resource usage to avoid the large latency and packet loss with the assistance of GNN, which leverages the connectivity structure of the THz UAV network.
Moreover, even if MADDPG realizes RE comparable to GLOVE, it cannot learn to mitigate the severe latency and packet loss.
Hence, GLOVE outperforms all the benchmark solutions in terms of lower latency while maintaining zero packet loss.

\begin{figure}[t] 
\centering
        \includegraphics[width=\linewidth]{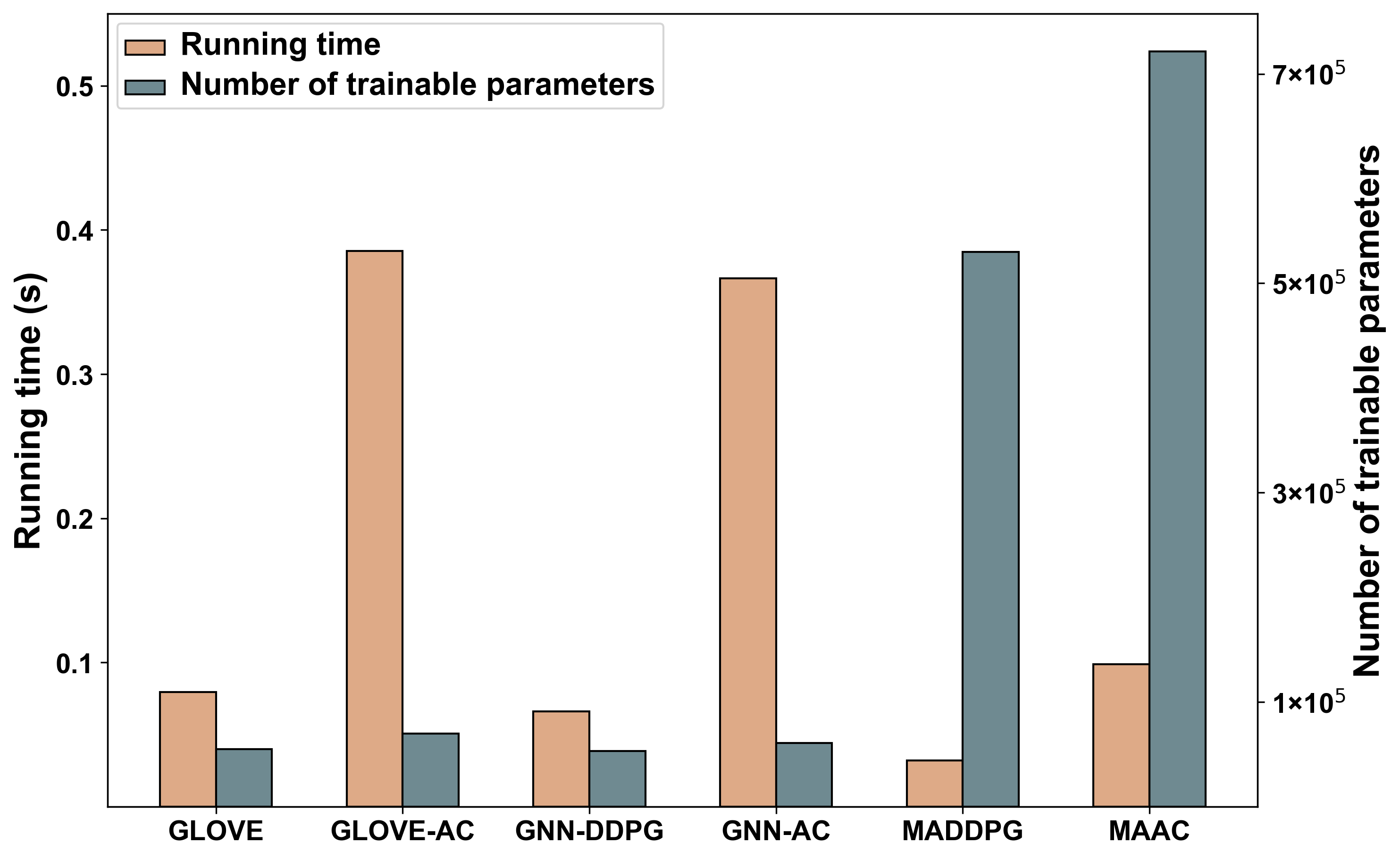} 
        \caption{Running time and number of trainable parameters comparisons.}
        \label{fig: running time and memory}
\end{figure}

On account of the realistic implementation of DRL algorithms, the running time and the number of trainable parameters are of critical importance as well.
The running time reflects the computational burden and the speed of DRL learning, while the number of training parameters is related to the memory requirement.
As illustrated in Fig.~\ref{fig: running time and memory}, GLOVE spends 79~ms for each training step with $5.5\times10^4$ trainable parameters.
Compared to GLOVE, GNN-DDPG uses 13~ms less running time and 2.6\% fewer training parameters.
Therefore, the mechanism putting an emphasis on self-features can substantially improve the performance of the GNN-aided DDPG algorithm (regarding RE, latency, as well as packet loss), without significantly compromising the running speed and memory efficiency.
By training multiple agents in a parallel manner, MADDPG realizes a running time that is 60\% shorter than GLOVE.
However, the multi-agent structure causes MADDPG to use 9.7 times as many trainable parameters as GLOVE.
Similarly, compared to GLOVE-AC, GNN-AC has a lower running time and fewer trainable parameters, while MAAC achieves a faster running speed at the expense of more trainable parameters.
Furthermore,  the discrete action selection processes of the AC-based algorithms require multiple output neurons for the selection of every output action value.
In contrast, the DDPG-based algorithms directly output each action value with only one neuron.
Therefore, GLOVE, GNN-DDPG, and MADDPG have fewer numbers of trainable parameters as well as substantially shorter running times than GLOVE-AC, GNN-AC, and MAAC, respectively, revealing higher computational efficiency and memory efficiency.

\section{Conclusion}
\label{conclusion}

In this paper, we proposed the GLOVE algorithm to jointly allocate power and sub-array resources in the dynamic THz UAV network with the target of RE maximization.
Particularly, on one hand, the proposed GLOVE algorithm utilizes GNN layers to learn the relationship among UAVs from the knowledge of the connectivity structure for the network topology in every time slot.
On the other hand, the important self-node features are also emphasized via FC layers.
A multi-task architecture is employed in GLOVE as well, enabling the cooperative training of power allocation and sub-array assignment for each UAV.
Additionally, GLOVE exploits safe initialization and safe exploration to prevent severe packet loss and latency throughout the training process.

Our experimental results validate that GLOVE realizes the lowest RE in the dynamic THz UAV network, compared to the benchmark methods.
In addition, during the entire training process, GLOVE achieves a maximal latency of 15~ms as well as zero packet loss, which are substantially lower than those of benchmark approaches (i.e., more than 46~ms maximal latency and more than $4\times10^5$ lost packets).
Moreover, the adopted structure that emphasizes the self-node features of each UAV improves RE, as well as mitigates latency and packet loss, in comparison with the GNN-aided DRL architecture, without considerably increasing computational and memory burdens.

	\bibliographystyle{IEEEtran}
	\bibliography{bibdata}

\begin{thebibliography}{10}
\providecommand{\url}[1]{#1}
\csname url@samestyle\endcsname
\providecommand{\newblock}{\relax}
\providecommand{\bibinfo}[2]{#2}
\providecommand{\BIBentrySTDinterwordspacing}{\spaceskip=0pt\relax}
\providecommand{\BIBentryALTinterwordstretchfactor}{4}
\providecommand{\BIBentryALTinterwordspacing}{\spaceskip=\fontdimen2\font plus
\BIBentryALTinterwordstretchfactor\fontdimen3\font minus \fontdimen4\font\relax}
\providecommand{\BIBforeignlanguage}[2]{{%
\expandafter\ifx\csname l@#1\endcsname\relax
\typeout{** WARNING: IEEEtran.bst: No hyphenation pattern has been}%
\typeout{** loaded for the language `#1'. Using the pattern for}%
\typeout{** the default language instead.}%
\else
\language=\csname l@#1\endcsname
\fi
#2}}
\providecommand{\BIBdecl}{\relax}
\BIBdecl

\bibitem{yan2019comprehensive}
C.~Yan, L.~Fu, J.~Zhang, and J.~Wang, ``A comprehensive survey on {UAV} communication channel modeling,'' \emph{IEEE Access}, vol.~7, pp. 107\,769--107\,792, 2019.

\bibitem{azari2022thz}
M.~M. Azari, S.~Solanki, S.~Chatzinotas, and M.~Bennis, ``{THz}-empowered {UAV}s in {6G}: Opportunities, challenges, and trade-offs,'' \emph{IEEE Communications Magazine}, vol.~60, no.~5, pp. 24--30, 2022.

\bibitem{kokkoniemi2021channel}
J.~Kokkoniemi, J.~M. Jornet, V.~Petrov, Y.~Koucheryavy, and M.~Juntti, ``Channel modeling and performance analysis of airplane-satellite {Terahertz} band communications,'' \emph{IEEE Transactions on Vehicular Technology}, vol.~70, no.~3, pp. 2047--2061, 2021.

\bibitem{gupta2015survey}
L.~Gupta, R.~Jain, and G.~Vaszkun, ``Survey of important issues in {UAV} communication networks,'' \emph{IEEE Communications Surveys \& Tutorials}, vol.~18, no.~2, pp. 1123--1152, 2015.

\bibitem{zeng2017energy}
Y.~Zeng and R.~Zhang, ``Energy-efficient {UAV} communication with trajectory optimization,'' \emph{IEEE Transactions on Wireless Communications}, vol.~16, no.~6, pp. 3747--3760, 2017.

\bibitem{akyildiz2022terahertz}
I.~F. Akyildiz, C.~Han, Z.~Hu, S.~Nie, and J.~M. Jornet, ``Terahertz band communication: An old problem revisited and research directions for the next decade,'' \emph{IEEE Transactions on Communications}, vol.~70, no.~6, pp. 4250--4285, 2022.

\bibitem{han2021hybrid}
C.~Han, L.~Yan, and J.~Yuan, ``Hybrid beamforming for {Terahertz} wireless communications: Challenges, architectures, and open problems,'' \emph{IEEE Wireless Communications}, vol.~28, no.~4, pp. 198--204, 2021.

\bibitem{meng2023uav}
K.~Meng, Q.~Wu, J.~Xu, W.~Chen, Z.~Feng, R.~Schober, and A.~L. Swindlehurst, ``{UAV}-enabled integrated sensing and communication: Opportunities and challenges,'' \emph{IEEE Wireless Communications}, vol.~31, no.~2, pp. 97--104, 2024.

\bibitem{hu2023deep}
Z.~Hu, C.~Han, and X.~Wang, ``Deep reinforcement learning based cross-layer design in {Terahertz} mesh backhaul networks,'' \emph{IEEE/ACM Transactions on Networking}, vol.~32, no.~3, pp. 2159--2173, 2024.

\bibitem{zhai2019antenna}
B.~Zhai, A.~Tang, C.~Huang, C.~Han, and X.~Wang, ``Antenna subarray management for hybrid beamforming in millimeter-wave mesh backhaul networks,'' \emph{Nano Communication Networks}, vol.~19, pp. 92--101, 2019.

\bibitem{zhang2024cooperative}
J.~Zhang, Y.~Wu, and M.~Zhou, ``Cooperative dual-task path planning for persistent surveillance and emergency handling by multiple unmanned ground vehicles,'' \emph{IEEE Transactions on Intelligent Transportation Systems}, vol.~25, no.~11, pp. 16\,288--16\,299, 2024.

\bibitem{jeong2022transport}
C.~Jeong, C.-J. Chun, W.-Y. Shin, and I.-M. Kim, ``Transport capacity optimization for resource allocation in {Tera-IoT} networks,'' \emph{IEEE Internet of Things Journal}, vol.~9, no.~16, pp. 15\,270--15\,284, 2022.

\bibitem{tong2022joint}
S.~Tong, Y.~Liu, J.~Mi{\v{s}}i{\'c}, X.~Chang, Z.~Zhang, and C.~Wang, ``Joint task offloading and resource allocation for fog-based intelligent transportation systems: A {UAV}-enabled multi-hop collaboration paradigm,'' \emph{IEEE Transactions on Intelligent Transportation Systems}, vol.~24, no.~11, pp. 12\,933--12\,948, 2022.

\bibitem{kim2020energy}
T.~Kim and D.~Qiao, ``Energy-efficient data collection for {IoT} networks via cooperative multi-hop {UAV} networks,'' \emph{IEEE Transactions on Vehicular Technology}, vol.~69, no.~11, pp. 13\,796--13\,811, 2020.

\bibitem{cui2019multi}
J.~Cui, Y.~Liu, and A.~Nallanathan, ``Multi-agent reinforcement learning-based resource allocation for {UAV} networks,'' \emph{IEEE Transactions on Wireless Communications}, vol.~19, no.~2, pp. 729--743, 2019.

\bibitem{zhou2022resource}
S.~Zhou, Y.~Cheng, X.~Lei, Q.~Peng, J.~Wang, and S.~Li, ``Resource allocation in {UAV}-assisted networks: A clustering-aided reinforcement learning approach,'' \emph{IEEE Transactions on Vehicular Technology}, vol.~71, no.~11, pp. 12\,088--12\,103, 2022.

\bibitem{bai2025dynamic}
Y.~Bai, B.~Xie, Y.~Liu, Z.~Chang, and R.~J{\"a}ntti, ``Dynamic {UAV} deployment in multi-{UAV} wireless networks: A multi-modal feature-based deep reinforcement learning approach,'' \emph{IEEE Internet of Things Journal}, early access, 2025.

\bibitem{park2023joint}
Y.~M. Park, S.~S. Hassan, Y.~K. Tun, Z.~Han, and C.~S. Hong, ``Joint trajectory and resource optimization of {MEC}-assisted {UAVs} in sub-{THz} networks: A resources-based multi-agent proximal policy optimization {DRL} with attention mechanism,'' \emph{IEEE Transactions on Vehicular Technology}, vol.~73, no.~2, pp. 2003--2016, 2023.

\bibitem{vestin2017low}
J.~Vestin and A.~Kassler, ``Low frequency assist for {mmWave} backhaul-the case for {SDN} resiliency mechanisms,'' in \emph{Proc. of IEEE ICC Workshops}, 2017, pp. 205--210.

\bibitem{fidler2014guide}
M.~Fidler and A.~Rizk, ``A guide to the stochastic network calculus,'' \emph{IEEE Communications Surveys \& Tutorials}, vol.~17, no.~1, pp. 92--105, 2014.

\bibitem{yan2020dynamic}
L.~Yan, C.~Han, and J.~Yuan, ``A dynamic array-of-subarrays architecture and hybrid precoding algorithms for {Terahertz} wireless communications,'' \emph{IEEE Journal on Selected Areas in Communications}, vol.~38, no.~9, pp. 2041--2056, 2020.

\bibitem{cao2007multihop}
M.~Cao, X.~Wang, S.-j. Kim, and M.~Madihian, ``Multi-hop wireless backhaul networks: a cross-layer design paradigm,'' \emph{IEEE Journal on Selected Areas in Communications}, vol.~25, no.~4, pp. 738--748, 2007.

\bibitem{walter2017time}
M.~Walter, D.~Shutin, and A.~Dammann, ``Time-variant{ Doppler PDFs} and characteristic functions for the vehicle-to-vehicle channel,'' \emph{IEEE Transactions on Vehicular Technology}, vol.~66, no.~12, pp. 10\,748--10\,763, 2017.

\bibitem{peng2021hybrid}
D.~Peng, A.~Bandi, Y.~Li, S.~Chatzinotas, and B.~Ottersten, ``Hybrid beamforming, user scheduling, and resource allocation for integrated terrestrial-satellite communication,'' \emph{IEEE Transactions on Vehicular Technology}, vol.~70, no.~9, pp. 8868--8882, 2021.

\bibitem{chang2021joint}
B.~Chang, X.~Yan, L.~Zhang, Z.~Chen, L.~Li, and M.~A. Imran, ``Joint communication and control for {mmWave/THz} beam alignment in {V2X} networks,'' \emph{IEEE Internet of Things Journal}, vol.~9, no.~13, pp. 11\,203--11\,213, 2021.

\bibitem{boulogeorgos2019analytical}
A.-A.~A. Boulogeorgos, E.~N. Papasotiriou, and A.~Alexiou, ``Analytical performance assessment of {THz} wireless systems,'' \emph{IEEE Access}, vol.~7, pp. 11\,436--11\,453, 2019.

\bibitem{yang2024universal}
Z.~Yang, W.~Gao, and C.~Han, ``A universal attenuation model of {Terahertz} wave in space-air-ground channel medium,'' \emph{IEEE Open Journal of the Communications Society}, vol.~5, pp. 2333--2342, 2024.

\bibitem{zhang2019joint}
X.~Zhang, C.~Han, and X.~Wang, ``Joint beamforming-power-bandwidth allocation in {Terahertz} {NOMA} networks,'' in \emph{Proc. of IEEE SECON}, 2019, pp. 1--9.

\bibitem{lei2020deep}
W.~Lei, Y.~Ye, and M.~Xiao, ``Deep reinforcement learning-based spectrum allocation in integrated access and backhaul networks,'' \emph{IEEE Transactions on Cognitive Communications and Networking}, vol.~6, no.~3, pp. 970--979, 2020.

\bibitem{tang2014resource}
J.~Tang, D.~K. So, E.~Alsusa, and K.~A. Hamdi, ``Resource efficiency: A new paradigm on energy efficiency and spectral efficiency tradeoff,'' \emph{IEEE Transactions on Wireless Communications}, vol.~13, no.~8, pp. 4656--4669, 2014.

\bibitem{liu2022deep}
Y.~Liu, J.~Yan, and X.~Zhao, ``Deep reinforcement learning based latency minimization for mobile edge computing with virtualization in maritime {UAV} communication network,'' \emph{IEEE Transactions on Vehicular Technology}, vol.~71, no.~4, pp. 4225--4236, 2022.

\bibitem{chen2021hybrid}
Y.~Chen, L.~Yan, and C.~Han, ``Hybrid spherical-and planar-wave modeling and {DCNN}-powered estimation of {Terahertz} ultra-massive {MIMO} channels,'' \emph{IEEE Transactions on Communications}, vol.~69, no.~10, pp. 7063--7076, 2021.

\bibitem{RLbook}
{Richard S. Sutton, Andrew G. Barto}, \emph{Reinforcement Learning: An Introduction}.\hskip 1em plus 0.5em minus 0.4em\relax {MIT Press}, 2018.

\bibitem{murti2022constrained}
F.~W. Murti, S.~Ali, and M.~Latva-Aho, ``Constrained deep reinforcement based functional split optimization in virtualized {RAN}s,'' \emph{IEEE Transactions on Wireless Communications}, vol.~21, no.~11, pp. 9850--9864, 2022.

\bibitem{marchesini2022exploring}
E.~Marchesini, D.~Corsi, and A.~Farinelli, ``Exploring safer behaviors for deep reinforcement learning,'' in \emph{Proc. of AAAI}, vol.~36, no.~7, 2022, pp. 7701--7709.

\bibitem{xu2021crpo}
T.~Xu, Y.~Liang, and G.~Lan, ``{CRPO}: A new approach for safe reinforcement learning with convergence guarantee,'' in \emph{Proc. of ICML}, 2021, pp. 11\,480--11\,491.

\bibitem{wu2020dynamic}
W.~Wu, N.~Chen, C.~Zhou, M.~Li, X.~Shen, W.~Zhuang, and X.~Li, ``Dynamic {RAN} slicing for service-oriented vehicular networks via constrained learning,'' \emph{IEEE Journal on Selected Areas in Communications}, vol.~39, no.~7, pp. 2076--2089, 2020.

\bibitem{lillicrap2015continuous}
T.~P. Lillicrap, J.~J. Hunt, A.~Pritzel, N.~Heess, T.~Erez, Y.~Tassa, D.~Silver, and D.~Wierstra, ``Continuous control with deep reinforcement learning,'' in \emph{Proc. of ICLR}, 2016, pp. 1--14.

\bibitem{lu2009simple}
P.~Lu and H.-C. Yang, ``A simple and efficient user-scheduling strategy for {RUB-based} multiuser {MIMO} systems and its sum-rate analysis,'' \emph{IEEE Transactions on Vehicular Technology}, vol.~58, no.~9, pp. 4860--4867, 2009.

\bibitem{you2020l2}
Y.~You, T.~Chen, Z.~Wang, and Y.~Shen, ``L2-{GCN}: Layer-wise and learned efficient training of graph convolutional networks,'' in \emph{Proc. of IEEE/CVF CVPR}, Jun. 2020, pp. 2127--2135.

\bibitem{hu2024multi}
Z.~Hu, C.~Han, Y.~Deng, and X.~Wang, ``Multi-task deep reinforcement learning for {Terahertz} {NOMA} resource allocation with hybrid discrete and continuous actions,'' \emph{IEEE Transactions on Vehicular Technology}, vol.~73, no.~8, pp. 11\,647--11\,663, 2024.

\bibitem{fawaz2018uav}
W.~Fawaz, C.~Abou-Rjeily, and C.~Assi, ``{UAV}-aided cooperation for {FSO} communication systems,'' \emph{IEEE Communications Magazine}, vol.~56, no.~1, pp. 70--75, 2018.

\bibitem{qin2024drl}
P.~Qin, Y.~Fu, J.~Zhang, S.~Geng, J.~Liu, and X.~Zhao, ``{DRL}-based resource allocation and trajectory planning for {NOMA}-enabled multi-{UAV} collaborative caching {6G} network,'' \emph{IEEE Transactions on Vehicular Technology}, vol.~73, no.~6, pp. 8750--8764, 2024.

\bibitem{xia2021multi}
Q.~Xia and J.~M. Jornet, ``Multi-hop relaying distribution strategies for {Terahertz}-band communication networks: A cross-layer analysis,'' \emph{IEEE Transactions on Wireless Communications}, vol.~21, no.~7, pp. 5075--5089, 2021.

\bibitem{deng2022resource}
Y.~Deng, H.~Jiang, P.~Cai, T.~Wu, P.~Zhou, B.~Li, H.~Lu, J.~Wu, X.~Chen, and K.~Wang, ``Resource provisioning for mitigating edge {DDoS} attacks in {MEC}-enabled {SDVN},'' \emph{IEEE Internet of Things Journal}, vol.~9, no.~23, pp. 24\,264--24\,280, 2022.

\bibitem{dong2021intelligent}
T.~Dong, Z.~Zhuang, Q.~Qi, J.~Wang, H.~Sun, F.~R. Yu, T.~Sun, C.~Zhou, and J.~Liao, ``Intelligent joint network slicing and routing via {GCN}-powered multi-task deep reinforcement learning,'' \emph{IEEE Transactions on Cognitive Communications and Networking}, vol.~8, no.~2, pp. 1269--1286, 2021.

\bibitem{gao2021game}
A.~Gao, Q.~Wang, W.~Liang, and Z.~Ding, ``Game combined multi-agent reinforcement learning approach for {UAV} assisted offloading,'' \emph{IEEE Transactions on Vehicular Technology}, vol.~70, no.~12, pp. 12\,888--12\,901, 2021.

\bibitem{araf2022uav}
S.~Araf, A.~S. Saha, S.~H. Kazi, N.~H. Tran, and M.~G.~R. Alam, ``{UAV} assisted cooperative caching on network edge using multi-agent actor-critic reinforcement learning,'' \emph{IEEE Transactions on Vehicular Technology}, vol.~72, no.~2, pp. 2322--2337, 2022.

\bibitem{iqbal2019actor}
S.~Iqbal and F.~Sha, ``Actor-attention-critic for multi-agent reinforcement learning,'' in \emph{Proc. of ICML}, 2019, pp. 2961--2970.

\end{thebibliography}
\end{document}